\pdfoutput=1

\documentclass[11pt,dvipsnames]{article}

\usepackage[acceptedWithA]{misc/tacl2018v2}

\usepackage[utf8]{inputenc} 
\usepackage[T1]{fontenc}    
\usepackage{times}          
\usepackage{inconsolata}    
\usepackage{pifont}

\usepackage{microtype}      
\usepackage{url}            
\usepackage{hyperref}       

\usepackage{rotating}
\usepackage{makecell}
\usepackage{booktabs}       
\usepackage{array}          
\usepackage{multirow}       
\usepackage{adjustbox}      
\usepackage{colortbl}       
\definecolor{Gray}{gray}{0.98}
\newcolumntype{g}{>{\columncolor{Gray}}c}
\usepackage{tabularx}

\usepackage{xcolor}

\usepackage{amsmath}        
\usepackage{amssymb}        
\usepackage{amsfonts}       
\usepackage{mathtools}      
\usepackage{amsthm}         
\usepackage{bm}
\usepackage{esvect}

\usepackage{cleveref}

\usepackage[inline, shortlabels]{enumitem} 

\usepackage{graphicx}       
\usepackage{subfigure}      
\usepackage{xspace}         

\usepackage[algoruled,ruled,vlined,noend]{algorithm2e} 

\usepackage{soul}           
\usepackage{changepage}     

\usepackage{todonotes}

\definecolor{TodoColor}{rgb}{1,0.7,0.6}
\definecolor{TodoColor2}{rgb}{0.7,0.7,0.9}
\definecolor{TodoColor3}{rgb}{0.5,0.8,0.5}
\definecolor{NoHighlightColor}{RGB}{57,74,245} 
\definecolor{UnsupervisedColor}{RGB}{230,149,10} 
\definecolor{SupervisedColor}{RGB}{220,13,152} 
\definecolor{OracleColor}{RGB}{28,154,14} 
\definecolor{RandomColor}{HTML}{888888}
\definecolor{TransColor}{HTML}{54278f}

\definecolor{MajorErrorColor}{HTML}{ed6a05}
\definecolor{MinorErrorColor}{RGB}{230, 187, 0}

\newcommand{\EnNl}{English$\rightarrow$Dutch\xspace}
\newcommand{\EnIt}{English$\rightarrow$Italian\xspace}
\newcommand{\EnDe}{English$\rightarrow$German\xspace}

\newcommand{\nohighlight}{\textcolor{NoHighlightColor}{\textbf{\textit{no highlight}}}\xspace}
\newcommand{\unsupervised}{\textcolor{UnsupervisedColor}{\textbf{\textit{unsupervised}}}\xspace}
\newcommand{\supervised}{\textcolor{SupervisedColor}{\textbf{\textit{supervised}}}\xspace}
\newcommand{\oracle}{\textcolor{OracleColor}{\textbf{\textit{oracle}}}\xspace}
\newcommand{\NoHighlight}{\textcolor{NoHighlightColor}{\textbf{No Highlight}}\xspace}
\newcommand{\Unsupervised}{\textcolor{UnsupervisedColor}{\textbf{Unsupervised}}\xspace}
\newcommand{\Supervised}{\textcolor{SupervisedColor}{\textbf{Supervised}}\xspace}
\newcommand{\Oracle}{\textcolor{OracleColor}{\textbf{Oracle}}\xspace}
\newcommand{\Random}{\textcolor{RandomColor}{\textbf{Random}}\xspace}

\newcommand{\NoHighlightShort}{\textcolor{NoHighlightColor}{\textbf{No High.}}\xspace}
\newcommand{\UnsupervisedShort}{\textcolor{UnsupervisedColor}{\textbf{Unsup.}}\xspace}
\newcommand{\SupervisedShort}{\textcolor{SupervisedColor}{\textbf{Sup.}}\xspace}
\newcommand{\OracleShort}{\textcolor{OracleColor}{\textbf{Oracle}}\xspace}

\newcommand{\notepos}[1]{$\textcolor{OliveGreen}{\null_{\uparrow#1}}$}
\newcommand{\noteneg}[1]{$\textcolor{BrickRed}{\null_{\downarrow#1}}$}

\newcommand{\nohighvv}[1]{{\color{NoHighlightColor}\vv{\color{black}{#1}}}}

\def\modblocks#1#2#3#4{
    \vspace{0.07cm}
    \hspace{-0.34cm}
    \begin{tikzpicture}[x=1em, y=1.3em]
        \draw [use as bounding box] (0,0) rectangle (4,1.8);
        \pgfmathsetmacro\NoHighlightTot{#1 * (1.8/5)}
        \pgfmathsetmacro\OracleTot{#2 * (1.8/5)}
        \pgfmathsetmacro\UnsupervisedTot{#3 * (1.8/5)}
        \pgfmathsetmacro\SupervisedTot{#4 * (1.8/5)}
        \fill [fill=NoHighlightColor]  (0,0) rectangle (1,\NoHighlightTot);
        \fill [fill=OracleColor]       (1,0) rectangle (2,\OracleTot);
        \fill [fill=UnsupervisedColor] (2,0) rectangle (3,\UnsupervisedTot);
        \fill [fill=SupervisedColor]   (3,0) rectangle (4,\SupervisedTot);
        \draw [thick, dashed] (0,0.9) -- (4,0.9);
    \end{tikzpicture}
    \hspace{-0.3cm}
    \vspace{-0.4cm}
}

\def\modblockshigh#1#2#3{
    \vspace{0.07cm}
    \hspace{-0.34cm}
    \begin{tikzpicture}[x=1em, y=1.3em]
        \draw [use as bounding box] (0,0) rectangle (4,1.8);
        \pgfmathsetmacro\OracleTot{#1 * (1.8/5)}
        \pgfmathsetmacro\UnsupervisedTot{#2 * (1.8/5)}
        \pgfmathsetmacro\SupervisedTot{#3 * (1.8/5)}
        \fill [fill=OracleColor]       (0,0) rectangle (1.33,\OracleTot);
        \fill [fill=UnsupervisedColor] (1.33,0) rectangle (2.66,\UnsupervisedTot);
        \fill [fill=SupervisedColor]   (2.66,0) rectangle (3.99,\SupervisedTot);
        \draw [thick, dashed] (0,0.9) -- (4,0.9);
    \end{tikzpicture}
    \hspace{-0.3cm}
    \vspace{-0.4cm}
}

\newcommand{\Both}{\footnotesize\textcolor{OliveGreen}{Both}}
\newcommand{\Neither}{\footnotesize\textcolor{BrickRed}{Neither}}
\newcommand{\NldOnly}{\textcolor{Apricot}{\textsc{nld}}}
\newcommand{\ItaOnly}{\textcolor{Apricot}{\textsc{ita}}}

\graphicspath{{img/}}

\title{QE4PE: Word-level Quality Estimation for Human Post-Editing}

\author{
  Gabriele Sarti$^1$ ~\;~ Vilém Zouhar$^2$ ~\;~ Grzegorz Chrupa\l{}a$^3$ \\ ~\;~ \textbf{Ana Guerberof-Arenas}$^1$ ~\;~ \textbf{Malvina Nissim}$^1$ ~\;~ \textbf{Arianna Bisazza}$^1$\vspace{3mm}\\
  $^1$CLCG, University of Groningen ~\;~ $^2$ETH Zürich ~\;~ $^3$CSAI, Tilburg University \vspace{3mm}\\
  $^1$\small{\texttt{\{g.sarti, a.guerberof.arenas, m.nissim, a.bisazza\}@rug.nl}}\\
  $^2$\small{\texttt{vzouhar@inf.ethz.ch}} ~\;~ $^3$\small{\texttt{grzegorz@chrupala.me}}
}

\begin{document}
\maketitle

\begin{abstract}
Word-level quality estimation (QE) methods aim to detect erroneous spans in machine translations, which can direct and facilitate human post-editing.
While the accuracy of word-level QE systems has been assessed extensively, their usability and downstream influence on the speed, quality and editing choices of human post-editing remain understudied.
In this study, we investigate the impact of word-level QE on machine translation (MT) post-editing in a realistic setting involving 42 professional post-editors across two translation directions.
We compare four error-span highlight modalities, including supervised and uncertainty-based word-level QE methods, for identifying potential errors in the outputs of a state-of-the-art neural MT model.
Post-editing effort and productivity are estimated from behavioral logs, while quality improvements are assessed by word- and segment-level human annotation.
We find that domain, language and editors' speed are critical factors in determining highlights' effectiveness, with modest differences between human-made and automated QE highlights underlining a gap between accuracy and usability in professional workflows.
\end{abstract}

\section{Introduction}
\label{sec:intro}

Recent years have seen a steady increase in the quality of machine translation (MT) systems and their widespread adoption in professional translation workflows \citep{kocmi-etal-2024-findings}.
Still, human post-editing of MT outputs remains a fundamental step to ensure high-quality translations, particularly for challenging textual domains requiring native fluency and specialized terminology \citep{liu-etal-2024-beyond-human}.
%
Quality estimation (QE) techniques were introduced to reduce post-editing effort by automatically identifying problematic MT outputs without the need for human-written reference translations and were quickly integrated into industry platforms~\citep{tamchyna-2021-deploying}.
\textit{Segment-level} QE models correlate well with human perception of quality~\citep{freitag-etal-2024-llms} and exceed the performance of reference-based metrics in specific settings~\citep{rei-etal-2021-references,amrhein-etal-2022-aces,amrhein-etal-2023-aces}.
On the other hand, \textit{word-level} QE methods for identifying error spans requiring revision have received less attention in the past due to their modest agreement with human annotations, despite their promise for more granular and interpretable quality assessment in line with modern MT practices~\citep{zerva-etal-2024-findings}.
In particular, while the accuracy of these approaches is regularly assessed in evaluation campaigns, research has rarely focused on assessing the impact of such techniques in realistic post-editing workflows, with notable exceptions suggesting limited benefits~\citep{shenoy-etal-2021-investigating,sugyeong-etal-2022-word}.
This hinders current QE evaluation practices: by foregoing experimental evaluation with human editors, it is implicitly assumed that word-level QE will become helpful once sufficient accuracy is reached, without accounting for the additional challenges towards a successful integration of these methods in post-editing workflows.

\begin{figure}[t]
\includegraphics[width=\linewidth]{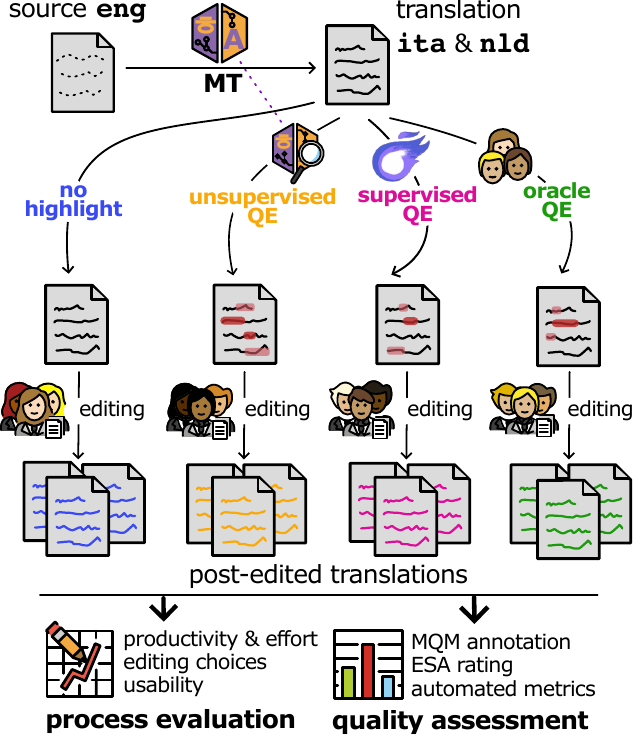}
\caption{A summary of the QE4PE study. Documents are translated by a neural MT model and reviewed by professional editors across two translation directions and four highlight modalities. Editing effort, productivity and usability across modalities are estimated from editing logs and questionnaires. Finally, the quality of MT and edited outputs is assessed with MQM/ESA human annotations and automatic metrics.}\vspace{-10pt}
\label{fig:highlevel_qe4pe}
\end{figure}

In this study, which we dub QE4PE (\textbf{Q}uality \textbf{E}stimation for \textbf{P}ost \textbf{E}diting), we address this gap by conducting a large-scale study with 42 professional translators for the \EnIt and \EnNl directions to measure the impact of word-level QE on editing quality, productivity and usability.
We aim for a realistic and reproducible setup, employing the high-quality open-source NLLB 3.3B MT model~\citep{nllb} to translate challenging documents from biomedical and social media domains. We then conduct a controlled evaluation of post-editing with error spans in four \textit{highlight modalities}, i.e. using highlights derived from four word-level QE methods:
a \supervised state-of-the-art QE model trained on human error annotations (XCOMET, \citealp{guerreiro-etal-2024-xcomet}), an \unsupervised method leveraging the uncertainty of the MT model during generation, \oracle error spans obtained from the consensus of previous human post-editors, and a \nohighlight baseline.
The human post-editing is performed using \textsc{GroTE}, a simple online interface we built to support the real-time logging of granular editing data, enabling a quantitative assessment of editing effort and productivity across highlight modalities.
We also survey professionals using an online questionnaire to collect qualitative feedback about the usability and quality of the MT model, as well as the interface and error span highlights.
Finally, a subset of the original MT outputs and their post-edited variants is annotated following the MQM and ESA protocols~\citep{lommel-2013-multidimensional,kocmi-etal-2024-error} to verify quality improvements after post-editing.
See \Cref{fig:highlevel_qe4pe} for an overview of the study.
Our work represents a step towards an evaluation of translation technologies that is centered on users' experience~\citep{guerberof-arenas-etal-2023-towards,savoldi-etal-2025-translation}.

We release all data, code and the \textsc{GroTE} editing interface to foster future studies on the usability of error span highlighting techniques for other word-level QE methods and translation directions.\footnote{Data: \href{https://huggingface.co/datasets/gsarti/qe4pe}{\texttt{hf.co/gsarti/qe4pe}}. Code: \href{https://github.com/gsarti/qe4pe}{\texttt{gsarti/qe4pe}}.} 

\section{Related Work}
\label{sec:related-work}

\paragraph{MT Post-editing}
Human post-editing of MT outputs is increasingly common in professional translator workflows, as it was shown to increase the productivity of translators while preserving translation quality across multiple domains \citep{liu-etal-2024-beyond-human}.
However, many factors were found to influence the variability of post-editing productivity across setups, including MT quality \citep{zouhar-etal-2021-neural}, interface familiarity \citep{laubli-etal-2021-impact}, individual variability and source-target languages typological similarity \citep{sarti-etal-2022-divemt}.
Studies evaluating the post-editing process generally focus on \textit{productivity}, i.e. number of processed words/characters per minute, and the temporal, technical and cognitive dimensions of post-editing \textit{effort}, operationalized through behavioral metrics such as editing time, keystrokes and pauses \citep{krings2001repairing,sarti-etal-2022-divemt}.
We adopt these metrics for the QE4PE study and relate them to different highlight modalities.

\paragraph{Quality estimation for MT}
The field of quality estimation was initially concerned with MT model uncertainty~\citep{blatz-etal-2004-confidence,specia-etal-2009-estimating}, but in time began focusing on predicting translation quality even without using references~(\citealp{turchi-etal-2013-coping,turchi-etal-2014-adaptive,kepler-etal-2019-openkiwi,thompson-post-2020-automatic} \textit{inter alia}). Advances in segment- and word-level QE research are regularly assessed in annual WMT campaigns~\citep{fomicheva-etal-2021-eval4nlp,zerva-etal-2022-findings,zerva-etal-2024-findings,blain-etal-2023-findings}, where the best-performing QE systems are usually Transformer-based language models trained on human quality judgments, such as the popular COMET model suite~\citep{rei-etal-2020-comet,rei-etal-2021-references,rei-etal-2022-comet}. The widespread adoption of the fine-grained Multidimensional Quality Metrics scale~(MQM, \citealp{lommel-2013-multidimensional}) prompted a paradigm shift in MT evaluation, leading to new QE metrics predicting quality at various granularity levels~\citep{kocmi-federmann-2023-gemba,fernandes-etal-2023-devil,guerreiro-etal-2024-xcomet}. Aside from supervised models, \textit{unsupervised} methods exploiting model uncertainty and its internal mechanisms were proposed as efficient alternatives to identify potential error spans in MT outputs (\citealp{fomicheva-etal-2020-unsupervised,dale-etal-2023-detecting,xu-etal-2023-understanding,himmi-etal-2024-enhanced}, surveyed by~\citealt{leiter-etal-2024-explainable}).
In this work, we compare the downstream effectiveness of state-of-the-art supervised and unsupervised word-level QE metrics for post-editing settings.

\paragraph{QE for Human Post-Editing Workflows}
Automatic QE methods are widely used in the translation industry for triaging automatic translations~\citep{tamchyna-2021-deploying}.
While QE usage has been found helpful to increase the confidence and speed of human assessment~\citep{mehandru-etal-2023-physician,zouhar-etal-2025-ai}, an incautious usage of these techniques can lead to a misplaced over-reliance on model predictions~\citep{zouhar-etal-2021-backtranslation}.
Interfaces supporting word-level error highlights were developed for studying MT post-editing~\citep{coppers2018intellingo,herbig-etal-2020-mmpe} and code reviewing~\citep{sun-etal-2022-investigating,vasconcelos-etal-2024-generation}, with results suggesting that striking the right balance of user-provided information is fundamental to improve the editing experience and prevent cognitive overload. Most similar to our study, \citet{shenoy-etal-2021-investigating} investigated the effect of synthetic word-level QE highlights for \EnDe post-editing on Wikipedia data, concluding that word-level QE accuracy was at the time still insufficient to produce tangible productivity benefits in human editing workflows.
In this work, we expand the scope of such evaluation by including two translation directions, two challenging real-world text domains and state-of-the-art MT and QE systems and methods.

\section{Experimental Setup}
\label{sec:experiment_setup}

\subsection{Structure of the Study}
\label{ssec:structure}

Our study is organized in five stages:

\paragraph{1) Oracle Post-editing} As a preliminary step, segments later used in the main assessment are post-edited by three professionals per direction using their preferred interface without logging. This allows us to obtain post-edits and produce \oracle word-level spans based on the editing consensus of multiple human professionals. Translators involved in this stage are not involved further in the study.

\paragraph{2) Pretask (\textsc{Pre})} The pretask allows the \textit{core translators} (12 per language direction, see \Cref{sec:participants}) to familiarize themselves with the \textsc{GroTE} interface and text highlights. Before starting, all translators complete a questionnaire to provide demographic and professional information about their profile (\Cref{tab:questionnaire-all}). In the pretask, all translators work in an identical setup, post-editing a small set of documents similar to those of the main task with \supervised highlights.
We assign core translators into three groups based on their speed from editing logs (4 translators per group for \textit{faster}, \textit{average} and \textit{slower} groups in each direction).
Individuals from each group are then assigned randomly to each highlight modality to ensure an equal representation of editing speeds, resulting in 1 \textit{faster}, 1 \textit{average} and 1 \textit{slower} translator for each highlight modality. This procedure is repeated independently for both translation directions.

\paragraph{3) Main Task (\textsc{Main})} This task, conducted in the two weeks following the pretask, covers the majority of the collected data and is the main object of study for the analyses of \Cref{sec:analysis}. In the main task, 24 core translators work on the same texts using the \textsc{GroTE} interface, with three translators per modality in each translation direction, as shown in \Cref{fig:highlevel_qe4pe}. After the main task, translators complete a questionnaire on the quality and usability of the MT outputs, the interface and, where applicable, word highlights.\footnote{We do not disclose the highlight modality to translators to avoid biasing their judgment in the evaluation.}

\paragraph{4) Post-Task (\textsc{Post})} After \textsc{Main}, the 12 core translators per direction are asked to post-edit an additional small set of related documents with \textsc{GroTE}, but this time working all with the \nohighlight modality. This step lets us obtain baseline editing patterns for each translator to estimate individual speed and editing differences across highlight modalities without the confounder of interface proficiency accounted for in the \textsc{Pre} stage.

\paragraph{5) Quality Assessment (QA)} Finally, a subset consisting of 148 main task segments is randomly selected for manual annotation by six new translators per direction (see \Cref{sec:participants}). For each segment, the original MT output and all its post-edited versions are annotated with MQM error spans, including minor/major error severity and a subset of MQM error categories including e.g. mistranslations, omissions and stylistic errors~\citep{lommel-2013-multidimensional}.\footnote{See \Cref{fig:qa} for an overview of setup and guidelines.} Moreover, the annotator proposes corrections for each error span, ultimately providing a 0-100 quality score matching the common DA scoring adopted in multiple WMT campaigns. We adopt this scoring system, which closely adheres to the ESA evaluation protocol~\citep{kocmi-etal-2024-error}, following recent results showing its effectiveness and efficiency for ranking MT system.

In summary, for each translation direction, we collect 3 full sets of oracle post-edits, 12 full sets of edits with behavioral logs for \textsc{pre}, \textsc{main} and \textsc{post} task data, and 13 subsets of main task data (12 post-edits, plus the original MT output) annotated with MQM error spans, corrections and segment-level ESA ratings. Moreover, we also collect 12 pre- and post-task questionnaire responses from \textit{core set} translators to obtain a qualitative view of the editing process.

\subsection{Highlight Modalities}
\label{ssec:modalities}

We conduct our study on four highlight modalities across two severity levels (\textit{minor} and \textit{major} errors).
Using multiple severity levels follows the current MT evaluation practices~\citep{freitag-etal-2021-experts,freitag-etal-2024-llms}, and previous results showing that users tend to prefer more granular and informative word-level highlights~\citep{shenoy-etal-2021-investigating,vasconcelos-etal-2024-generation}. The highlight modalities we employ are:

\paragraph{\NoHighlight} The text is presented as-is, without any highlighted spans. This setting serves as a baseline to estimate the default post-editing quality and productivity using our interface.

\paragraph{\Oracle} Following the Oracle Post-editing phase, we produce oracle error spans from the editing consensus of human post-editors. We label text spans that were edited by two out of three translators as \textit{minor}, and those edited by all three translators as \textit{major}, following the intuition that more critical errors are more likely to be identified by several annotators, while minor changes will show more variance across subjects. This modality serves as a best-case scenario, providing an upper bound for future improvements in word-level QE quality. 

\paragraph{\Supervised} In this setting, word-level error spans are obtained using XCOMET-XXL \citep{guerreiro-etal-2024-xcomet}, which is a multilingual Transformer encoder \citep{xlmr} further trained for joint word- and sentence-level QE prediction.
We select XCOMET-XXL in light of its broad adoption, open accessibility and state-of-the-art performance in QE across several translation directions \citep{zerva-etal-2024-findings}.
For the severity levels, we use the labels predicted by the model, mapping \textit{critical} labels to the \textit{major} level.

\begin{table}[t]
    \centering
    \footnotesize
    \begin{tabular}{lcccc}
        \toprule[1.5pt]
        \multirow{2}{*}{\bf Method} & \multicolumn{2}{c}{\bf DivEMT} & \multicolumn{2}{c}{\bf QE4PE} \\
         \cmidrule(lr){2-3}
         \cmidrule(lr){4-5}
         & \bf \textsc{en}-\textsc{it} & \bf \textsc{en}-\textsc{nl} & \bf  \textsc{en}-\textsc{it} & \bf \textsc{en}-\textsc{nl} \\
        \midrule
        Logprobs    & 0.18 & 0.19 & 0.10 & 0.09 \\
        MCD Var.    & \textbf{0.41} & \textbf{0.42} & \textbf{0.23} & \textbf{0.31} \\
        XCOMET (\SupervisedShort)     & 0.34 & 0.35 & 0.16 & 0.19 \\
        \midrule
        Avg. Trans. & -   & -   & 0.53 & 0.55\\
        \bottomrule[1.5pt]
    \end{tabular}
    \caption{Average Precision (AP) between metrics and reference error spans.}
    \vspace{-3pt}
\label{tab:unsup-perf}
\end{table}

\paragraph{\Unsupervised} In this modality, we exploit the access to the MT model producing the original translations to obtain \textit{uncertainty-based highlights}. As a preliminary evaluation to select a capable unsupervised word-level QE method, we evaluate two unsupervised QE methods employing token log-probabilities assigned by MT model to predict human post-edits: raw negative log-probabilities (Logprobs), corresponding to the surprisal assigned by the MT model to every generated token, and their variance for 10 steps of Monte Carlo Dropout (MCD Var., \citealp{mcdropout}). We employ surprisal-based metrics following previous work showing their effectiveness in predicting translation errors~\citep{fomicheva-specia-2019-taking} and human editing time~\citep{lim-etal-2024-predicting}. We collect scores for the \EnIt and \EnNl directions of QE4PE Oracle post-edits and DivEMT~\citep{sarti-etal-2022-divemt} to identify the best-performing method, using metric scores extracted from the original models used for translation to predict human post-edits. We use average precision (AP) as a threshold-agnostic performance metric for the tested continuous methods. \Oracle highlights obtained from the consensus of three annotator in the first stage of the study are used as reference for QE4PE, while a single set of post-edits is available for DivEMT. The XCOMET-XXL model used for \Supervised highlights, and the average agreement of individual \Oracle editors with the consensus label are also included for comparison.~\Cref{tab:unsup-perf}\footnote{Full results in \Cref{tab:unsup_selection}. Highlights are extended from tokens to words to match the granularity of other modalities.} show a strong performance for the MCD Var. method, even surpassing the accuracy of the supervised XCOMET model across both datasets. Hence, we select MCD Var. for the \Unsupervised highlight modality, setting value thresholds for minor/major errors to match the respective highlighted word proportions in the \Supervised modality to ensure a fair comparison.

\subsection{Data and MT model}
\label{sec:data}

\paragraph{MT Model}
On the one hand, the MT model must achieve \textit{high translation quality} in the selected languages to ensure our experimental setup applies to state-of-the-art proprietary systems.
Still, the MT model should be \textit{open-source} and have a \textit{manageable size} to ensure reproducible findings and enable the computation of uncertainty for the unsupervised setting.
All considered, we use NLLB 3.3B \citep{nllb}, a widely-used MT model achieving industry-level performances across 200 languages~\citep{moslem-etal-2023-adaptive}.

\begin{table}[t]
    \centering
    \footnotesize
    \begin{tabular}{llccc}
        \toprule[1.5pt]
         \bf Task & \bf Domain & \bf \# Docs & \bf \# Seg. & \bf \# Words \\
        \midrule
        \multirow{2}{*}{\textsc{Pre}} &  Social & 4 & 23 & 539 \\
        & Biomed. & 2 & 15 & 348 \\
        \midrule
        \multirow{2}{*}{\textsc{Main}} & Social & 30 & 160 & 3375 \\
        & Biomed. & 21 & 165 & 3384 \\
        \midrule
        \multirow{2}{*}{\textsc{Post}} &  Social & 6 & 34 & 841 \\
        & Biomed. & 2 & 16 & 257 \\
        \midrule
        \multicolumn{2}{c}{\bf Total} & 64 & 413 & 8744 \\
        \bottomrule[1.5pt]
    \end{tabular}
    \caption{Statistics for QE4PE data.}
    \vspace{-3pt}
\label{tab:data_stats}
\end{table}

\paragraph{Data selection}
We begin by selecting two translation directions, \EnIt and \EnNl, according to the availability of professional translators from our industrial partners. We intentionally focus on out-of-English translations as they are generally more challenging for modern MT models \citep{kocmi-etal-2023-findings}. We aim to identify documents that are manageable for professional translators without domain-specific expertise but still prove challenging for our MT model to ensure a sufficient amount of error spans across modalities. Since original references for our selected translation direction were not available, we do not have a direct mean to compare MT quality in the two languages. However, according to our human MQM assessment in \Cref{sec:quality} (\Cref{tab:mqm-errors-main-min-maj}), NLLB produces a comparable amount of errors across Dutch and Italian MT, suggesting similar quality.

We begin by translating 3,672 multi-segment English documents from the WMT23 General and Biomedical MT shared tasks~\citep{kocmi-etal-2023-findings,neves-etal-2023-findings} and MT test suites to Dutch and Italian. Our choice for these specialized domains, as opposed to e.g. generic news articles, is driven by the real-world needs of the translation industry for domain-specific post-editing support~\citep{eschbach-dymanus-etal-2024-exploring,li-etal-2025-transbench}. Moreover, focusing on domains that are considerably more challenging for MT systems than news, as shown by recent WMT campaigns~\citep{neves-etal-2024-findings}, ensures a sufficient amount of MT errors to support a sound comparison of word-level QE methods. 
Then, XCOMET-XXL is used to produce a first set of segment-level QE scores and word-level error spans for all segments.
To make the study tractable, we further narrow down the selection of documents according to several heuristics to ensure a realistic editing experience and a balanced occurrence of error spans (details in \Cref{app:data_stats}). This procedure yields 351 documents, from which we manually select a subset of 64 documents (413 segments, 8,744 source words per post-editor) across two domains:

\begin{table}
    \footnotesize
    \centering
    \setlength{\tabcolsep}{4.5pt}
    \begin{tabular}{lp{6cm}}
        \toprule[1.5pt]
        Source\textsubscript{~\textsc{en}} & So why is it that people jump through extra hoops to install Google Maps? \\
        \midrule
        \NoHighlightShort & Quindi perché le persone devono fare un salto in più per installare Google Maps? \\
        \OracleShort & \textcolor{MinorErrorColor}{Quindi} perché le persone \textcolor{MajorErrorColor}{devono fare} \textcolor{MajorErrorColor}{un salto in più} per installare Google Maps?\\
        \SupervisedShort & \textcolor{MinorErrorColor}{Quindi perché} le persone devono \textcolor{MajorErrorColor}{fare un salto in più} per installare Google Maps?\\
        \UnsupervisedShort & Quindi perché le persone \textcolor{MajorErrorColor}{devono} \textcolor{MinorErrorColor}{fare} un \textcolor{MajorErrorColor}{salto} \textcolor{MinorErrorColor}{in} più per installare Google Maps?\\
        \midrule
        PE\textsubscript{\NoHighlightShort} & Quindi perché le persone devono fare un \textcolor{YellowGreen}{passaggio} in più per installare Google Maps?\\
        PE\textsubscript{\OracleShort} & \textcolor{YellowGreen}{Allora,} perché le persone \textcolor{YellowGreen}{fanno} un \textcolor{YellowGreen}{passaggio} in più per installare Google Maps?\\
        PE\textsubscript{\SupervisedShort} & Quindi perché le persone \textcolor{YellowGreen}{fanno passaggi} in più per installare Google Maps? \\
        PE\textsubscript{\UnsupervisedShort} & Quindi perché le persone \textcolor{YellowGreen}{fanno i salti mortali} per installare Google Maps? \\
        \bottomrule[1.5pt]
    \end{tabular}
    \caption{\textsc{en}$\rightarrow$\textsc{it} example from the QE4PE dataset, showing \textcolor{MinorErrorColor}{minor}/\textcolor{MajorErrorColor}{major} word highlights and a single post-edit per modality, with modified words \textcolor{YellowGreen}{highlighted}.}
    \label{tab:example-highlights-edits}
    \vspace{-5pt}
\end{table}

\begin{itemize}[itemsep=1mm,left=0mm,topsep=1mm]
\item \textbf{Social media posts}, including Mastodon posts from the WMT23 General Task~\citep{kocmi-etal-2023-findings} English$\leftrightarrow$German evaluation and Reddit comments from the Robustness Challenge Set for Machine Translation (RoCS-MT; \citealp{bawden-sagot-2023-rocs}), displaying atypical language use, such as slang or acronymization. 
\item \textbf{Biomedical abstracts} extracted from PubMed from the WMT23 Biomedical Translation Task~\citep{neves-etal-2023-findings}, including domain-specific terminology. 
\end{itemize}

\noindent
\Cref{tab:data_stats} present statistics for the \textsc{Pre}, \textsc{Main} and \textsc{Post} editing stages, and \Cref{tab:example-highlights-edits} shows an example of highlights and edits. While the presence of multiple domains in the same task can render our post-editing setup less realistic, we deem it essential to test the cross-domain validity of our findings.

\paragraph{Critical Errors} Before producing highlights, we manually introduce 13 critical errors in main task segments to assess post-editing thoroughness. Errors are produced, for example, by negating statements, inverting the polarity of adjectives, inverting numbers, and corrupting acronyms. We replicate the errors in both translation directions to enable direct comparison. Most of these errors were correctly identified across all three highlight modalities (examples in~\Cref{tab:critical-errors-examples}).

\subsection{Participants}
\label{sec:participants}

For both directions, professional translation companies Translated Srl\footnote{\url{https://translated.com}} and Global Textware\footnote{\url{https://globaltextware.nl/}} recruited three translators for the Oracle post-editing stage, the core set of 12 translators working on \textsc{Pre}, \textsc{Main} and \textsc{Post} tasks, and six more translators for the QA stage, for a total of 21 translators per direction.
All translators were freelancers with native proficiency in their target language and self-assessed proficiency of at least C1 in English. Almost all translators had more than two years of professional translation experience and regularly post-edited MT outputs (details in \Cref{tab:questionnaire-all}).

\subsection{Editing Interface} 
\label{sec:interface}

\begin{figure}
\center
\includegraphics[width=\linewidth]{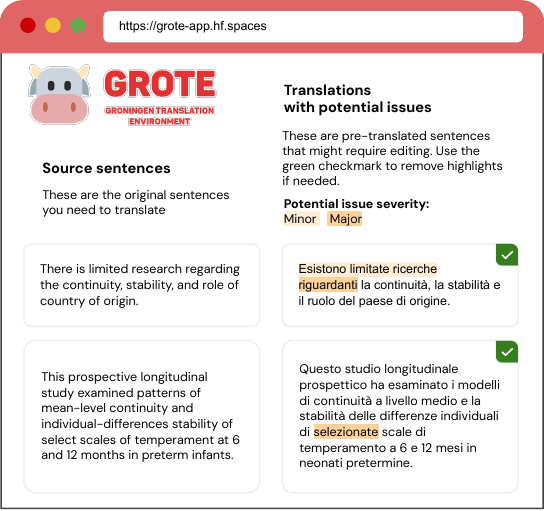}
\caption{An example of the QE4PE \textsc{GroTE} setup for two segments in an \EnIt document.}
\label{fig:grote_interface}
\vspace{-5pt}
\end{figure}

We develop a custom interface, which we name \textbf{Gro}ningen \textbf{T}ranslation \textbf{E}nvironment (\textsc{GroTE}, \Cref{fig:grote_interface}), to support editing over texts with word-level highlights. While the MMPE tool used by~\citet{shenoy-etal-2021-investigating} provide extensive multimodal functionalities~\citep{herbig-etal-2020-mmpe}, we aim for a bare-bones setup to avoid confounders in the evaluation. \textsc{GroTE} is a web interface based on Gradio~\citep{abid2019gradio} and hosted on the \href{https://hf.co/spaces}{Hugging Face Spaces} to enable multi-user data collection online.
Upon loading a document, source texts and MT outputs for all segments are presented in two columns following standard industry practices.
For modalities with highlights, the interface provides an informative message and supports the removal of all highlights on a segment via a button, with highlights on words disappearing automatically upon editing, as in~\citet{shenoy-etal-2021-investigating}. The interface supports real-time logging of user actions, allowing for the analysis of the editing process. In particular, we log the start/end times for each document, the accessing and exiting of segment textboxes, highlight removals, and keystrokes.

\textsc{GroTE} intentionally lacks standard features such as translation memories, glossaries, and spellchecking to ensure equal familiarity among translators, ultimately controlling for editor proficiency with these tools, as done in previous studies~\citep{shenoy-etal-2021-investigating,sarti-etal-2022-divemt}. While most translators noted the lack of features in our usability assessment, the majority also found the interface easy to set up, access, and use (\Cref{tab:questionnaire-all}).

\section{Analysis}
\label{sec:analysis}

\subsection{Productivity}
\label{sec:productivity}

\begin{figure}[t]
\includegraphics[width=\linewidth]{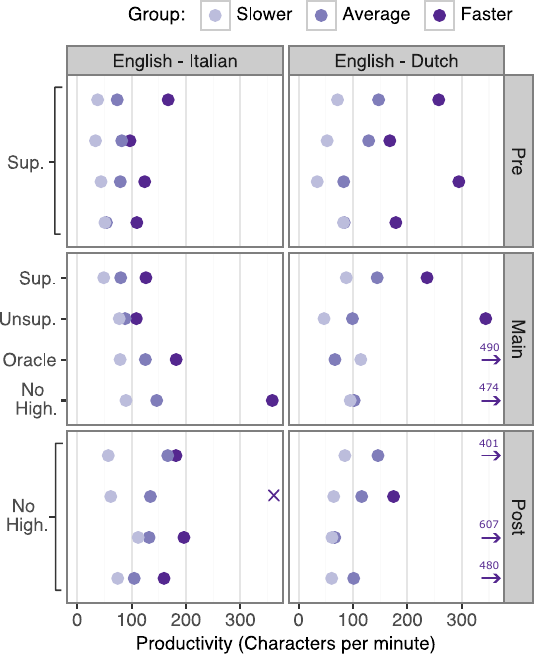}

\vspace{-3pt}
\caption{Productivity of post-editors across QE4PE stages (\textsc{Pre}, \textsc{Main}, \textsc{Post}).
The \textcolor{TransColor}{\ding{212}} marks outstanding entries and \textcolor{TransColor}{\ding{53}} marks missing data.
Each row corresponds to the same three translators across all stages.}\vspace{-5pt}
\label{fig:productivity}
\end{figure}

We obtain segment- and document-level edit times and compute editing \textit{productivity} as the number of processed source characters over the sum of all document-level edit times, measured in characters per minute.
To account for potential breaks taken by post-editors during editing, we filter out pauses between logged actions longer than 5 minutes. We note that this procedure does not significantly impact the overall ranking of translators, while ensuring a more robust evaluation of editing time.

\paragraph{Do Highlights Make Post-editors Faster?} \Cref{fig:productivity} shows translators' productivity across stages, with every dot corresponding to the productivity of a single individual. We observe that no highlight modality leads to systematically faster editing across all speed groups and that the ordering of \textsc{Pre}-task speed groups is maintained in the following stages despite the different highlight modalities. These results suggest that individual variability in editing speed is more critical than highlight modality in predicting editing speed. 
However, faster \EnNl translators achieve outstanding productivity, i.e.~$>2$ standard deviations above the overall mean ($>300$ char/min, \textcolor{TransColor}{\ding{212}} in~\Cref{fig:productivity}) almost exclusively in \NoHighlight, and, \Oracle~modalities, suggesting that lower-quality highlights hinder editing speed.

We validate these observations by fitting a negative binomial mixed-effect model on segment-level editing times (model details in \Cref{tab:edit_time_model}).
Excluding random factors such as translator and segment identity from the model produces a significant drop in explained variance, confirming the inherent variability of editing times ($R^2 = 0.93 \rightarrow 0.41$).
Model coefficients show that MT output length and the proportion of highlighted characters are the main factors driving an increase in editing times, possibly reflecting an increase in cognitive effort to process additional information.
We find highlights to have a significant impact on increasing the editing speed of \EnIt translators $(p < 0.001)$, but a minimal impact for \EnNl.
Comparing the productivity of the same translator editing with and without highlights (\textsc{Main} vs \textsc{Post}), two-thirds of the translators editing with highlights were up to two times slower on biomedical texts. However, the same proportion of translators was up to three times faster on social media texts across both directions.

In summary, we find that \textbf{highlight modalities are not predictive of edit times on their own}, but translation direction and domain play an important role in determining the effect of highlights on editing productivity.
We attribute these results to two main factors, which will remain central in the analysis of the following sections: (1) the different \textit{propensity of translators to act upon highlighted issues} in the two tested directions, and (2) the different \textit{nature of errors highlighted across domains}.

\subsection{Highlights and Edits}
\label{sec:highlights-edits}

We then examine how highlights are distributed across modalities and how they influence the editing choices of human post-editors.

\begin{table*}[ht]
    \setlength{\tabcolsep}{4.5pt}
    \small
    \centering
    \begin{tabular}{lcc|clcl|llll}
        \toprule[1.5pt]
        & \multicolumn{2}{c}{\textbf{Base Freq.}} &  \multicolumn{4}{c}{\textbf{Measured}} & \multicolumn{4}{c}{\textbf{Projected}}\\
        \cmidrule(lr){2-3}
        \cmidrule(lr){4-7}
        \cmidrule(lr){8-11}
        & $P(H)$& $P(E)$  & $P(E|H)$ & $\Lambda_E$ & $P(H|E)$ & $\Lambda_H$ & $\nohighvv{P}(E|H)$ & \multicolumn{1}{c}{$\nohighvv{\Lambda}_E$} & $\nohighvv{P}(H|E)$ & \multicolumn{1}{c}{$\nohighvv{\Lambda}_H$} \\
        \midrule
        \rowcolor{lightgray} \multicolumn{11}{c}{\textbf{\EnIt}}   \\
        \NoHighlightShort  & -    & 0.05 & -    & -    & -    & -   & -    & -    & -    & -    \\
        \Random            & 0.16 & -    & -    & -    & -    & -   & 0.06 & 1.20 & 0.18 & 1.20 \\
        \OracleShort       & 0.15 & 0.12 & \textbf{0.37} & \textbf{4.62} & \underline{\textbf{0.45}} & \textbf{4.1} & \textbf{0.18}\noteneg{0.19} & \textbf{6.00}\notepos{1.38} & \underline{\textbf{0.55}}\notepos{0.10} & \textbf{4.23}\notepos{0.14} \\
        \UnsupervisedShort & 0.16 & 0.13 & 0.25 & 2.27 & 0.21 & 2.2 & 0.11\noteneg{0.14} & 2.75\notepos{0.48} & 0.37\notepos{0.16} & 2.47\notepos{0.26} \\
        \SupervisedShort   & 0.12 & 0.16 & 0.28 & 2.00 & 0.22 & 2.0 & 0.14\noteneg{0.14} & 3.50\notepos{1.50} & 0.35\notepos{0.13} & 3.18\notepos{1.18} \\
        \midrule
        \rowcolor{lightgray} \multicolumn{11}{c}{\textbf{\EnNl}}   \\
        \NoHighlightShort  & -    & 0.14 & -    & -    & -    & -   & -    & -    & -    & -    \\
        \Random            & 0.17 & -    & -    & -    & -    & -    & 0.16 & 1.14 & 0.19 & 1.19 \\
        \OracleShort       & 0.20 & 0.10 & \textbf{0.26} & \underline{\textbf{4.33}} & \underline{\textbf{0.53}} & 3.12 & \textbf{0.28}\notepos{0.02} & \textbf{2.55}\noteneg{1.78} & \underline{\textbf{0.40}}\noteneg{0.13} & 2.35\noteneg{0.77} \\
        \UnsupervisedShort & 0.20 & 0.11 & 0.20 & 2.50 & 0.36 & 2.00 & 0.22\notepos{0.02} & 1.83\noteneg{0.67} & 0.31\noteneg{0.05} & 1.72\noteneg{0.28} \\
        \SupervisedShort   & 0.12 & 0.09 & 0.24 & 3.43 & 0.33 & \textbf{3.30} & \textbf{0.28}\notepos{0.04} & 2.33\noteneg{1.10} & 0.24\noteneg{0.09} & \textbf{2.40}\noteneg{0.90} \\
        \bottomrule[1.5pt]
    \end{tabular}
    \caption{Highlighting ($H$) and editing ($E$) average statistics across directions and highlight modalities. \textbf{Measured}: actual edits performed in the specified modality. \textbf{Projected}: using modality highlights over \NoHighlight~edits to account for editing biases~(\Cref{sec:proj-highlights}). Random highlights matching average word frequencies are used as \Random~baseline, and Projected \textcolor{OliveGreen}{increases}\notepos{}~/~\textcolor{BrickRed}{decreases}\noteneg{} compared to Measured counterparts are shown. Significant~\OracleShort~gains over all other modalities are \underline{underlined} ($p<0.05$ with Bonferroni correction).}
    \label{tab:highlights-edits}\vspace{-2pt}
\end{table*}

\paragraph{Agreement Across Modalities} First, we quantify how different modalities agree in terms of highlight distribution and editing.
We find that highlight overlaps across modalities range between 15\% and 39\% when comparing highlight modalities in a pairwise fashion, with the highest overlap for \EnIt social media and \EnNl biomedical texts.\footnote{Scores are normalized to account for highlight frequencies across modalities. Agreement is shown in \Cref{tab:highlight_agreement}.} Despite the relatively low highlight agreement, we find an average agreement of 73\% for post-edited characters across modalities.
This suggests that edits are generally uniform regardless of highlight modalities and are not necessarily restricted to highlighted spans.\footnote{Editing agreement is shown in \Cref{fig:edit_agreement}.}

\paragraph{Do Highlights Accurately Identify Potential Issues?} \Cref{tab:highlights-edits} (Base Freq.) shows raw highlight and edit frequencies across modalities. We observe different trends across the two language pairs: for \EnIt, post-editors working with highlights edit more than twice as much as translators with \NoHighlight, regardless of the highlight modality. On the contrary, for \EnNl they edit 33\% less in the same setting.
These results suggest a different attitude towards acting upon highlighted potential issues across the two translation directions, with \EnIt translators appearing to be conditioned to edit more when highlights are present.
We introduce four metrics to quantify highlights-edits overlap:

\begin{itemize}[itemsep=0.3mm,left=0mm,topsep=1mm]
\item $P(E|H)$ and $P(H|E)$, reflecting highlights' \textit{precision} and \textit{recall} in predicting edits, respectively.
\item $\Lambda_E\,{\stackrel{\text{\tiny def}}{=}}\,P(E|H)/P(E|\neg H)$ shows how much more likely an edit is to fall within rather than outside highlighted characters.
\item $\Lambda_H\,{\stackrel{\text{\tiny def}}{=}}\,P(H|E)/P(H|\neg E)$ shows how much more likely it is for a highlight to mark edited rather than unmodified spans.
\end{itemize}

\noindent
Intuitively, character-level recall $P(H|E)$ should be more indicative of highlight quality compared to precision $P(E|H)$, provided that word-level highlights can be useful even when not minimal.\footnote{For example, if the fully-highlighted word \textit{tradutt\underline{ore}} is changed to its feminine version \textit{tradutt\underline{rice}}, $P(H|E) = 1$ (edit correctly and fully predicted) but $P(E|H) = 0.3$ since word stem characters are left unchanged.}
\Cref{tab:highlights-edits} (Measured) shows metric values across the three highlight modalities (breakdowns by domain and speed shown in \Cref{tab:edit_highlights_stats_domain_modality,tab:edit_highlights_stats_domain_speed}).
As expected, \Oracle~highlights obtain the best performance in terms of precision and recall, with $P(H|E)$, in particular, being significantly higher than the other two modalities across both directions.

Surprisingly, \textbf{we find no significant precision and recall differences between \Supervised{} and \Unsupervised{} highlights}, despite the word-level QE training of XCOMET used in the former modality. Moreover, they support the potential of unsupervised, model internals-based techniques to complement or substitute more expensive supervised approaches. Still, likelihood ratios $\Lambda_E, \Lambda_H \gg  1$ for all modalities and directions indicate that highlights are 2-4 times more likely to precisely and comprehensively encompass edits than non-highlighted texts. This suggests that even imperfect highlights that do not reach~\Oracle-level quality might effectively direct editing efforts toward potential issues. We validate these observations by fitting a zero-inflated negative binomial mixed-effects model to predict segment-level edit rates.
Results confirm a significantly higher edit rate for \EnIt highlighted modalities and the social media domain with $p<0.001$ (features and significances shown in Appendix \Cref{tab:edit_rate_model}). We find a significant zero inflation associated with translator identity, suggesting the choice of leaving MT outputs unedited is highly subjective.

\paragraph{Do Highlights Influence Editing Choices?} \label{sec:proj-highlights}
Since in~\Cref{sec:productivity} we found the proportion of highlighted characters to impact the editing rate of translators, we question whether the relatively high $P(E|H)$ and $P(H|E)$ values might be artificially inflated by the eagerness of translators to intervene on highlighted spans. In other words, \textit{do highlights identify actual issues, or do they condition translators to edit when they otherwise would not?} To answer this, we propose to \textit{project} highlights from a selected modality---in which highlights were shown during editing---onto the edits performed by the \NoHighlight~translators on the same segments.
The resulting difference between measured and projected metrics can then be taken as an estimate for the impact of highlight presentation on their resulting accuracy.

To further ensure the soundness of our analysis, we use a set of projected \Random~highlights as a lower bound for highlight performance. To make the comparison fair, \Random~highlights are created by randomly highlighting words in MT outputs matching the average word-level highlight frequency across all highlighted modalities given the current domain and translation direction.
\Cref{tab:highlights-edits} (Projected) shows results for the three highlighted modalities. First, all projected metrics remain consistently above the \Random~baseline, suggesting a higher-than-chance ability to identify errors even for worst-performing highlight modalities. Projected precision scores $\nohighvv{P}(E|H)$ depend on edit frequency, and hence see a major decrease for \EnIt, where the \NoHighlight~edit rate $P(E)$ is much lower. However, the increase in $\nohighvv{\Lambda}_E$ across all \EnIt modalities confirms that, despite the lower edit proportion, highlighted texts remain notably more likely to be edited than non-highlighted ones.
Conversely, the lower $\nohighvv{\Lambda}_E$, $\nohighvv{P}(H|E)$ and $\nohighvv{\Lambda}_H$ for \EnNl show that edits become much less skewed towards highlighted spans in this direction when accounting for presentation bias. 

Overall, while the presence of highlights makes \EnIt translators more likely to intervene in MT outputs, their location in the MT output often pinpoints issues that would be edited regardless of highlighting. \EnNl translators, on the contrary, intervene at roughly the same rate regardless of highlights presence, but their edits are focused mainly on highlighted spans when they are present. This difference is consistent across all subjects in the two directions despite the identical setup and comparable MT and QE quality across languages. This suggests that cultural factors might play a non-trivial role in determining the usability and influence of QE methods regardless of span accuracy, a phenomenon previously observed in human-AI interaction studies~\citep{ge-etal-2024-culture}.

\begin{figure*}[t]
\centering
\includegraphics[width=0.9\textwidth]{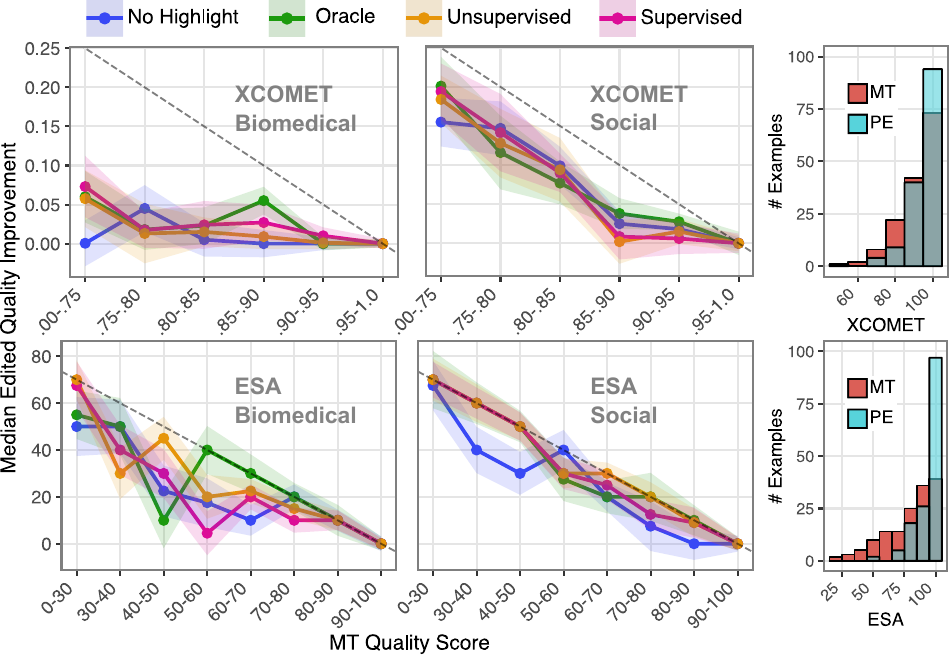}

\vspace{-4pt}
\caption{Median quality improvement for post-edited segments at various initial MT quality levels across domains and highlight modalities. Quality scores are estimated using XCOMET segment-level QE (top) and professional ESA annotations (bottom). Histograms show example counts across quality bins for the two metrics. Dotted lines show upper bounds for quality improvements given starting MT quality.}
\vspace{-5pt}
\label{fig:quality}
\end{figure*}

\subsection{Quality Assessment}
\label{sec:quality}

We continue our assessment by inspecting the quality of MT and post-edited outputs along three dimensions.
First, we use XCOMET segment-level QE ratings as an automatic approximation of quality and compare them to human-annotated quality scores collected in the last phase of our study. For efficiency, these are obtained for the 0-100 Direct Assessment scale commonly used in QE evaluation~\citep{specia-etal-2020-findings-wmt}, but following an initial step of MQM error annotation to condition scoring on found errors, as prescribed by the ESA protocol~\citep{kocmi-etal-2024-error}. Then, MQM error span annotations are used to analyze the distribution of error categories. Finally, we manually assess critical errors, which were inserted to quantify highlight modalities effect on unambiguous issues.

\paragraph{Do Highlights Influence Post-Editing Quality?}
In this stage, we focus particularly on  \textit{edited quality improvements}, i.e.\ how post-editing the same MT outputs under different highlight conditions influences the resulting quality of translations. We operationalize this assessment using human ratings and automatic metrics to score MT and post-edited translations, using their difference as the effective quality gain after the post-editing stage. Scores for this metric are generally positive, i.e. human post-editing improves quality, and bounded by the maximal achievable quality gain given the initial MT quality.
\Cref{fig:quality} shows median improvement values across quality bins defined from the distribution of initial MT quality scores (shown in histograms), in which all post-edited versions of each MT output appear as separate observations. Positive median scores confirm that post-edits generally lead to quality improvements across all tested settings. However, we observe different trends across the two metrics: across both domains, XCOMET greatly underestimates the human-assessed ESA quality improvement, especially for biomedical texts where it shows negligible improvement regardless of the initial MT quality.
These results echo recent findings cautioning users against the poor performance of trained MT metrics for unseen domains and high-quality translations~\citep{agrawal-etal-2024-automatic-metrics,zouhar-etal-2024-fine}.
Focusing on the more reliable ESA scores, we observe large quality improvements from post-editing, as shown by near-maximal quality gains across most bins and highlight modalities.
While \NoHighlight seems to underperform other modalities in the social media domain, the lack of more notable differences in gains across highlight modalities suggests that \textbf{highlights' quality impact might not be evident in terms of segment-level quality}, motivating our next steps in the quality analysis.

We also find no clear relationship between translator speed and edited quality improvements, suggesting that higher productivity does not come at a cost for faster translators (\Cref{fig:quality_across_speed}). This finding confirms that neglecting errors is not the cause of different editing patterns from previous sections.

\begin{table}[t]
    \setlength{\tabcolsep}{3.5pt}
    \footnotesize
    \centering
    \begin{tabular}{ll|ccccc}
    \toprule[1.5pt]
 & & \textbf{MT} & \textbf{\NoHighlightShort} & \textbf{\OracleShort} & \textbf{\UnsupervisedShort} & \textbf{\SupervisedShort} \\
\midrule
\multirow{4}{*}{\rotatebox[origin=c]{90}{\textbf{Italian}}} & Acc.
                 & 26 / 31 & 12 / 17 & \textbf{6  / 12} & 30 / 22 & 22 / 24 \\
& Style          & 17 / 33 & 5  / 33 & \textbf{0  / 15} & 5  / 35 & 4  / 31 \\
& Ling.          & 12 / 31 & 2  / 29 & \textbf{0  / 11} & 7  / 20 & 2  / 16 \\
\cmidrule(lr){2-7}
& \textbf{Tot.}  & 55 / 95 & 19 / 79 & \textbf{6  / 38} & 42 / 77 & 28 / 71 \\
\midrule
\midrule
\multirow{4}{*}{\rotatebox[origin=c]{90}{\textbf{Dutch}}} & Acc.
                 & 32 / 40 & 20 / \textbf{31} & 26 / 37 & \textbf{14} / 39 & 20 / 39 \\
& Style          & 3  / 29 & \textbf{1}  / 28 & \textbf{1}  / 23 & 2  / \textbf{18} & 7  / 50 \\
& Ling.          & 4  / 26 & 3  / 18 & 5  / 28 & \textbf{2  / 9}  & 3  / 15 \\
\cmidrule(lr){2-7}
& \textbf{Tot.}  & 39 / 95 & 24 / 77 & 32 / 88 & \textbf{18 / 66} & 30 / 104\\
\bottomrule[1.5pt]
\end{tabular}
\caption{Minor / major MQM error counts averaged across $n = 3$ translators per highlight modality for every translation direction on the QA \textsc{Main} subset. Lowest minor / major error counts per language are \textbf{bolded}.}
\label{tab:mqm-errors-main-min-maj}
\vspace{-12pt}
\end{table}

\paragraph{Which Error Types Do Highlights Identify?} \Cref{tab:mqm-errors-main-min-maj} shows a breakdown of MQM annotations for MT and all highlight modalities using the \textit{Accuracy}, \textit{Style} and \textit{Linguistic} macro-categories of MQM errors.\footnote{Full micro-category breakdown in \Cref{tab:critical-errors}, per-domain breakdown in \Cref{fig:quality_mqm}. Category descriptions in \Cref{fig:qa}.} At this granularity, differences across modalities become visible, with overall error counts showing a clear relation to $\nohighvv{\Lambda}_E$ from \Cref{tab:highlights-edits} (\Oracle being remarkably better for~\EnIt, with milder and more uniform trends in \EnNl). At least for \EnIt, these results confirm that an observable quality improvement from editing with highlights is present in the best-case~\Oracle~scenario. By contrast, for \EnNl the \Unsupervised method is found to outperform even the \Oracle setting in reducing the amount of errors, while it fares relatively poorly for \EnIt. We also note a different distribution of Accuracy and Style errors, with the former being more common in biomedical texts while the latter appearing more often for translated social media posts (\Cref{fig:quality_mqm}). We posit that differences in error types across domains might explain the opposite productivity trends observed in \Cref{sec:productivity}: while highlighted accuracy errors might lead to time-consuming terminology verification in biomedical texts, style errors might be corrected more quickly and naturally in the social media domain.

\paragraph{Do Highlights Detect Critical Errors?} We examine whether the critical errors we inserted were detected by different modalities, finding that while most modalities fare decently with more than 62\% of critical errors highlighted,~\Unsupervised~is the only setting for which all errors are correctly highlighted across both directions. Then, critical errors are manually verified in all outputs, finding that 16-20\% more critical errors are edited in highlighted modalities compared to \NoHighlight (full results in \Cref{tab:critical-errors}).
Hence, \textbf{highlights might lead to narrow but tangible quality improvements that can go undetected in coarse quality assessments}, and finer-grained evaluations might be needed to quantify future improvements in word-level QE.

\subsection{Usability}

In post-task questionnaire answers (\Cref{tab:questionnaire}), most translators stated that MT outputs had average-to-high quality and that provided texts were challenging to translate. Highlights were generally found decently accurate, but they were generally not found useful to improve either productivity or quality (including \Oracle ones). Interestingly, despite the convincing gains for critical errors measured in the last section, most translators stated that highlights did not influence their editing and did not help them identify errors that would have otherwise been missed. Concretely, this suggests that the potential quality improvements might not be easily perceived by translators and might have secondary importance compared to the extra cognitive load elicited by highlighted spans. When asked to comment about highlights, several translators called them \textit{``more of an eye distraction, as they often weren’t actual mistakes''} and \textit{``not quite accurate enough to rely on them as a suggestion''}. Some translators also stated that missed errors led them to \textit{``disregarding the highlights to focus on checking each sentence''}. Despite their high quality, only one editor working with~\Oracle~highlights found highlights helpful in~\textit{``making the editing process faster and somehow easier''}. Taken together, these comments convincingly point to a negative perception of the quality and usefulness of highlights, suggesting that \textbf{improvement in QE accuracy may not be sufficient to improve QE usefulness} in editors' eyes.

\begin{table}[t]
    \center
    \footnotesize
    \begin{tabular}{>{\raggedright}p{4.1cm}m{1.3cm}m{1.3cm}}
    \toprule[1.5pt]
    \textbf{Question} &\hspace{-0.35cm}\textbf{Italian}\hspace{-0.2cm} & \hspace{-0.35cm}\textbf{Dutch}\hspace{-0.2cm} \\
    \midrule
    MT outputs were generally of high quality. & \modblocks{3.33}{3.66}{3}{3} & \modblocks{2.33}{3}{2.33}{3.66} \\
    Provided texts were challenging to translate. & \modblocks{3.33}{3.67}{3.67}{4.33} & \modblocks{4.67}{3.67}{4.33}{3.00} \\
    \midrule
    \textbf{Highlights ...} & & \\
    ... were generally accurate in detecting potential issues. & \modblockshigh{2.67}{2.67}{2} & \modblockshigh{3}{3}{2.67} \\
    ... were generally useful during editing. & \modblockshigh{2.33}{2.33}{1.67} & \modblockshigh{1.67}{3}{2.33} \\
    ... improved my editing productivity. & \modblockshigh{2.33}{2.33}{2.00} & \modblockshigh{2.0}{2.67}{1.67} \\
    ... improved the quality of my translations. & \modblockshigh{1.33}{2.67}{1.33} & \modblockshigh{2}{3.33}{1.67} \\
    ... required additional editing effort on my part. & \modblockshigh{2.00}{2.67}{2.67} & \modblockshigh{2.00}{2.67}{2.67} \\
    ... influenced my editing choices. & \modblockshigh{1}{2.00}{2.67} & \modblockshigh{1.67}{2.00}{2.00} \\
    ... helped identify errors I'd have otherwise missed. & \modblockshigh{2}{2.33}{3} & \modblockshigh{1.33}{2}{1.67} \\
    \bottomrule[1.5pt]
    \end{tabular}
    \caption{Post-task questionnaire responses. Bars represent responses ranging from 1--Strongly disagree (no bar) to 5--Strongly agree (full bar), averaged across $n=3$ translators per language for \NoHighlight, \Oracle, \Unsupervised, and \Supervised. Dotted line mark avg. judgments of 3--Neither agree nor disagree.}
    \label{tab:questionnaire}
    \vspace{-12pt}
\end{table}

\section{Conclusion}

This study evaluated the impact of various error-span highlighting modalities, including automatic and human-made ones, on the productivity and quality of human post-editing in a realistic professional setting. Our findings highlight the importance of domain, language and editors' speed in determining highlights' effect on productivity and quality, underscoring the need for broad evaluations encompassing diverse settings. The limited gains of human-made highlights over automatic QE and their indistinguishable perception from editors' assessment indicate that further gains in the accuracy of these techniques might not be the determining factor in improving their integration into post-editing workflows. In particular, future work might explore other directions to further assess and improve the usability of word-level QE highlights, for example, studying their impact on non-professional translators and language learners or combining them with edit suggestions to justify the presence of error spans.

\section{Limitations}
\label{sec:limitations}

Our study presents certain limitations that warrant consideration when interpreting its findings and for guiding future research.

Firstly, while we included two domains and translation directions to improve the generalizability of our findings, our results suggest that language and domain play an important role in defining the effectiveness of word-level QE for human post-editing. While we observed mild gains from word-level QE on our tested mid-resourced translation directions (\EnIt and \EnNl), we expect limited, if any, benefit of such approaches in low-resource languages and domains for which MT systems and QE methods are likely to underperform~\citep{sarti-etal-2022-divemt,zouhar-etal-2024-fine}. Furthermore, the domains tested in our study (biomedical and social media posts) provided concrete challenges in the form of specialized terminology and idiomatic expressions, respectively, which are known to hinder the quality of MT outputs~\citep{neves-etal-2024-findings,bawden-sagot-2023-rocs}. While future work should ensure our findings can be extended to other domains and languages, the limited benefits brought by the tested word-level QE methods in challenging settings suggest a limited usefulness for higher-resource languages and more standard domains such as news or Wiki texts.

Secondly, we acknowledge that several design choices in our evaluation setup, rather than pertaining to the QE methods themselves, may have influenced our results. These include, for instance, the specific procedure for discretizing continuous scores from the \Unsupervised method into error spans, and the method of obtaining \oracle highlights via majority voting among post-editors. While we believe these choices are justified within the context of our study, their impact on the outcomes cannot be entirely discounted. Future studies might benefit from a more fine-grained assessment of how such low-level decisions influence the perceived accuracy and usability of word-level QE.

Finally, subjective factors such as the translators' inherent propensity to edit, their prior opinions on the role of MT in post-editing, and their individual editing styles inevitably influenced both quantitative and qualitative assessments in this study. Although we attempted to mitigate these effects by ensuring a controlled evaluation setup for all professional translators and by using averaged judgments for translators working on the same highlight modality, we acknowledge that subjectivity might limit the reproducibility of our findings.

\section{Broader Impact and Ethical Considerations}
Our study explicitly centers the experience of professional translators, responding to recent calls for
user-centered evaluation of translation technologies. By prioritizing translators’ perspectives and productivity, we aim to contribute to methods that complement rather than replace human expertise.
Our findings highlight a gap between user perception and measured quality improvements, suggesting that future efforts should focus primarily on improving the usability of these methods in editing interfaces. In particular, new assistive approaches for post-editing should not only strive to increase productivity but rather reduce the cognitive burden associated with post-editing work. This insight is crucial for designing more user-centered quality estimation tools that genuinely support human work. Ultimately, our results suggest that subjective norms across different domains and cultures play an important role in determining the effectiveness of proposed methodologies, underscoring the importance of accounting for human factors when designing such evaluations. All participants in this study were professional translators who provided informed consent. The research protocol ensured anonymity and voluntary participation, with translators recruited and remunerated through professional translation providers. The released materials further promote transparency, enabling other researchers to reproduce and build upon our findings.

\section*{Acknowledgements}

Authors acknowledge the support of Imminent, which funded the data collection through a \href{https://imminent.translated.com/apply-for-your-grants}{ Language Technology Grant}. Authors are grateful to Translated S.r.l. and Global Textware B.v. and to all translators involved in this study for agreeing to participate in the data collection process. We thank TACL reviewers and editors for their insightful comments. Gabriele Sarti, Grzegorz Chrupała and Arianna Bisazza also acknowledge the support of the Dutch Research Council (NWO) for the project InDeep (NWA.1292.19.399). Arianna Bisazza is supported by the NWO Talent Program (VI.Vidi.221C.009). Ana Guerberof Arenas is supported by the ERC Grant CoG INCREC 101086819.

\bibliography{anthology.min.bib,custom.bib}

\begin{thebibliography}{63}
\expandafter\ifx\csname natexlab\endcsname\relax\def\natexlab#1{#1}\fi

\bibitem[{Abid et~al.(2019)Abid, Abdalla, Abid, Khan, Alfozan, and Zou}]{abid2019gradio}
Abubakar Abid, Ali Abdalla, Ali Abid, Dawood Khan, Abdulrahman Alfozan, and James Zou. 2019.
\newblock \href {https://arxiv.org/abs/1906.02569} {Gradio: Hassle-free sharing and testing of {ML} models in the wild}.
\newblock \emph{CoRR}, cs.LG/1906.02569v1.

\bibitem[{Agrawal et~al.(2024)Agrawal, Farinhas, Rei, and Martins}]{agrawal-etal-2024-automatic-metrics}
Sweta Agrawal, Ant{\'o}nio Farinhas, Ricardo Rei, and Andre Martins. 2024.
\newblock \href {https://doi.org/10.18653/v1/2024.emnlp-main.802} {Can automatic metrics assess high-quality translations?}
\newblock In \emph{Proceedings of the 2024 Conference on Empirical Methods in Natural Language Processing}, pages 14491--14502. Association for Computational Linguistics.

\bibitem[{Amrhein et~al.(2022)Amrhein, Moghe, and Guillou}]{amrhein-etal-2022-aces}
Chantal Amrhein, Nikita Moghe, and Liane Guillou. 2022.
\newblock \href {https://aclanthology.org/2022.wmt-1.44/} {{ACES}: Translation accuracy challenge sets for evaluating machine translation metrics}.
\newblock In \emph{Proceedings of the Seventh Conference on Machine Translation (WMT)}, pages 479--513. Association for Computational Linguistics.

\bibitem[{Amrhein et~al.(2023)Amrhein, Moghe, and Guillou}]{amrhein-etal-2023-aces}
Chantal Amrhein, Nikita Moghe, and Liane Guillou. 2023.
\newblock \href {https://doi.org/10.18653/v1/2023.wmt-1.57} {{ACES}: Translation accuracy challenge sets at {WMT} 2023}.
\newblock In \emph{Proceedings of the Eighth Conference on Machine Translation}, pages 695--712. Association for Computational Linguistics.

\bibitem[{Bawden and Sagot(2023)}]{bawden-sagot-2023-rocs}
Rachel Bawden and Beno{\^i}t Sagot. 2023.
\newblock \href {https://doi.org/10.18653/v1/2023.wmt-1.21} {{R}o{CS}-{MT}: Robustness challenge set for machine translation}.
\newblock In \emph{Proceedings of the Eighth Conference on Machine Translation}, pages 198--216. Association for Computational Linguistics.

\bibitem[{Blain et~al.(2023)Blain, Zerva, Rei, Guerreiro, Kanojia, C.~de Souza, Silva, Vaz, Jingxuan, Azadi, Orasan, and Martins}]{blain-etal-2023-findings}
Frederic Blain, Chrysoula Zerva, Ricardo Rei, Nuno~M. Guerreiro, Diptesh Kanojia, Jos{\'e}~G. C.~de Souza, Beatriz Silva, T{\^a}nia Vaz, Yan Jingxuan, Fatemeh Azadi, Constantin Orasan, and Andr{\'e} Martins. 2023.
\newblock \href {https://doi.org/10.18653/v1/2023.wmt-1.52} {Findings of the {WMT} 2023 shared task on quality estimation}.
\newblock In \emph{Proceedings of the Eighth Conference on Machine Translation}, pages 629--653. Association for Computational Linguistics.

\bibitem[{Blatz et~al.(2004)Blatz, Fitzgerald, Foster, Gandrabur, Goutte, Kulesza, Sanchis, and Ueffing}]{blatz-etal-2004-confidence}
John Blatz, Erin Fitzgerald, George Foster, Simona Gandrabur, Cyril Goutte, Alex Kulesza, Alberto Sanchis, and Nicola Ueffing. 2004.
\newblock \href {https://aclanthology.org/C04-1046/} {Confidence estimation for machine translation}.
\newblock In \emph{{COLING} 2004: Proceedings of the 20th International Conference on Computational Linguistics}, pages 315--321. COLING.

\bibitem[{Coppers et~al.(2018)Coppers, Van~den Bergh, Luyten, Coninx, Van~der Lek-Ciudin, Vanallemeersch, and Vandeghinste}]{coppers2018intellingo}
Sven Coppers, Jan Van~den Bergh, Kris Luyten, Karin Coninx, Iulianna Van~der Lek-Ciudin, Tom Vanallemeersch, and Vincent Vandeghinste. 2018.
\newblock \href {https://dl.acm.org/doi/abs/10.1145/3173574.3174098} {Intellingo: {An} intelligible translation environment}.
\newblock In \emph{Proceedings of the 2018 CHI Conference on Human Factors in Computing Systems}, pages 1--13.

\bibitem[{Dale et~al.(2023)Dale, Voita, Barrault, and Costa-juss{\`a}}]{dale-etal-2023-detecting}
David Dale, Elena Voita, Loic Barrault, and Marta~R. Costa-juss{\`a}. 2023.
\newblock \href {https://doi.org/10.18653/v1/2023.acl-long.3} {Detecting and mitigating hallucinations in machine translation: Model internal workings alone do well, sentence similarity {E}ven better}.
\newblock In \emph{Proceedings of the 61st Annual Meeting of the Association for Computational Linguistics (Volume 1: Long Papers)}, pages 36--50. Association for Computational Linguistics.

\bibitem[{Eo et~al.(2022)Eo, Park, Moon, Seo, and Lim}]{sugyeong-etal-2022-word}
Sugyeong Eo, Chanjun Park, Hyeonseok Moon, Jaehyung Seo, and Heuiseok Lim. 2022.
\newblock \href {https://doi.org/10.1109/ACCESS.2022.3169155} {Word-level quality estimation for korean-english neural machine translation}.
\newblock \emph{IEEE Access}, 10:44964--44973.

\bibitem[{Eschbach-Dymanus et~al.(2024)Eschbach-Dymanus, Essenberger, Buschbeck, and Exel}]{eschbach-dymanus-etal-2024-exploring}
Johannes Eschbach-Dymanus, Frank Essenberger, Bianka Buschbeck, and Miriam Exel. 2024.
\newblock \href {https://aclanthology.org/2024.eamt-1.51/} {Exploring the effectiveness of {LLM} domain adaptation for business {IT} machine translation}.
\newblock In \emph{Proceedings of the 25th Annual Conference of the European Association for Machine Translation (Volume 1)}, pages 610--622. European Association for Machine Translation (EAMT).

\bibitem[{Fernandes et~al.(2023)Fernandes, Deutsch, Finkelstein, Riley, Martins, Neubig, Garg, Clark, Freitag, and Firat}]{fernandes-etal-2023-devil}
Patrick Fernandes, Daniel Deutsch, Mara Finkelstein, Parker Riley, Andr{\'e} Martins, Graham Neubig, Ankush Garg, Jonathan Clark, Markus Freitag, and Orhan Firat. 2023.
\newblock \href {https://doi.org/10.18653/v1/2023.wmt-1.100} {The devil is in the errors: Leveraging large language models for fine-grained machine translation evaluation}.
\newblock In \emph{Proceedings of the Eighth Conference on Machine Translation}, pages 1066--1083. Association for Computational Linguistics.

\bibitem[{Fomicheva et~al.(2021)Fomicheva, Lertvittayakumjorn, Zhao, Eger, and Gao}]{fomicheva-etal-2021-eval4nlp}
Marina Fomicheva, Piyawat Lertvittayakumjorn, Wei Zhao, Steffen Eger, and Yang Gao. 2021.
\newblock \href {https://doi.org/10.18653/v1/2021.eval4nlp-1.17} {The {E}val4{NLP} shared task on explainable quality estimation: Overview and results}.
\newblock In \emph{Proceedings of the 2nd Workshop on Evaluation and Comparison of NLP Systems}, pages 165--178. Association for Computational Linguistics.

\bibitem[{Fomicheva and Specia(2019)}]{fomicheva-specia-2019-taking}
Marina Fomicheva and Lucia Specia. 2019.
\newblock \href {https://doi.org/10.1162/coli\_a\_00356} {Taking {MT} evaluation metrics to extremes: Beyond correlation with human judgments}.
\newblock \emph{Computational Linguistics}, 45(3):515--558.

\bibitem[{Fomicheva et~al.(2020)Fomicheva, Sun, Yankovskaya, Blain, Guzm{\'a}n, Fishel, Aletras, Chaudhary, and Specia}]{fomicheva-etal-2020-unsupervised}
Marina Fomicheva, Shuo Sun, Lisa Yankovskaya, Fr{\'e}d{\'e}ric Blain, Francisco Guzm{\'a}n, Mark Fishel, Nikolaos Aletras, Vishrav Chaudhary, and Lucia Specia. 2020.
\newblock \href {https://doi.org/10.1162/tacl\_a\_00330} {Unsupervised quality estimation for neural machine translation}.
\newblock \emph{Transactions of the Association for Computational Linguistics}, 8:539--555.

\bibitem[{Freitag et~al.(2021)Freitag, Foster, Grangier, Ratnakar, Tan, and Macherey}]{freitag-etal-2021-experts}
Markus Freitag, George Foster, David Grangier, Viresh Ratnakar, Qijun Tan, and Wolfgang Macherey. 2021.
\newblock \href {https://doi.org/10.1162/tacl\_a\_00437} {Experts, errors, and context: A large-scale study of human evaluation for machine translation}.
\newblock \emph{Transactions of the Association for Computational Linguistics}, 9:1460--1474.

\bibitem[{Freitag et~al.(2024)Freitag, Mathur, Deutsch, Lo, Avramidis, Rei, Thompson, Blain, Kocmi, Wang, Adelani, Buchicchio, Zerva, and Lavie}]{freitag-etal-2024-llms}
Markus Freitag, Nitika Mathur, Daniel Deutsch, Chi-Kiu Lo, Eleftherios Avramidis, Ricardo Rei, Brian Thompson, Frederic Blain, Tom Kocmi, Jiayi Wang, David~Ifeoluwa Adelani, Marianna Buchicchio, Chrysoula Zerva, and Alon Lavie. 2024.
\newblock \href {https://doi.org/10.18653/v1/2024.wmt-1.2} {Are {LLM}s breaking {MT} metrics? results of the {WMT}24 metrics shared task}.
\newblock In \emph{Proceedings of the Ninth Conference on Machine Translation}, pages 47--81. Association for Computational Linguistics.

\bibitem[{Gal and Ghahramani(2016)}]{mcdropout}
Yarin Gal and Zoubin Ghahramani. 2016.
\newblock \href {https://proceedings.mlr.press/v48/gal16.html} {Dropout as a bayesian approximation: {Representing} model uncertainty in deep learning}.
\newblock In \emph{Proceedings of The 33rd International Conference on Machine Learning}, volume~48 of \emph{Proceedings of Machine Learning Research}, pages 1050--1059, New York, New York, USA. PMLR.

\bibitem[{Ge et~al.(2024)Ge, Xu, Misaki, Markus, and Tsai}]{ge-etal-2024-culture}
Xiao Ge, Chunchen Xu, Daigo Misaki, Hazel~Rose Markus, and Jeanne~L Tsai. 2024.
\newblock \href {https://doi.org/10.1145/3613904.3642660} {How culture shapes what people want from ai}.
\newblock In \emph{Proceedings of the 2024 CHI Conference on Human Factors in Computing Systems}, CHI '24, New York, NY, USA. Association for Computing Machinery.

\bibitem[{Goyal et~al.(2021)Goyal, Du, Ott, Anantharaman, and Conneau}]{xlmr}
Naman Goyal, Jingfei Du, Myle Ott, Giri Anantharaman, and Alexis Conneau. 2021.
\newblock \href {https://arxiv.org/abs/2105.00572} {Larger-scale transformers for multilingual masked language modeling}.
\newblock \emph{CoRR}, cs.CL/2105.00572v1.

\bibitem[{Guerberof-Arenas and Moorkens(2023)}]{guerberof-arenas-etal-2023-towards}
Ana Guerberof-Arenas and Joss Moorkens. 2023.
\newblock \href {https://doi.org/10.1007/978-3-031-14689-3_7} {Ethics and machine translation: The end user perspective}.
\newblock In Helena Moniz and Carla Parra~Escart\'in, editors, \emph{Towards Responsible Machine Translation: Ethical and Legal Considerations in Machine Translation}, pages 113--133. Springer International Publishing, Cham.

\bibitem[{Guerreiro et~al.(2024)Guerreiro, Rei, Stigt, Coheur, Colombo, and Martins}]{guerreiro-etal-2024-xcomet}
Nuno~M. Guerreiro, Ricardo Rei, Daan~van Stigt, Luisa Coheur, Pierre Colombo, and Andr{\'e} F.~T. Martins. 2024.
\newblock \href {https://doi.org/10.1162/tacl\_a\_00683} {xcomet: Transparent machine translation evaluation through fine-grained error detection}.
\newblock \emph{Transactions of the Association for Computational Linguistics}, 12:979--995.

\bibitem[{Herbig et~al.(2020)Herbig, D{\"u}wel, Pal, Meladaki, Monshizadeh, Kr{\"u}ger, and van Genabith}]{herbig-etal-2020-mmpe}
Nico Herbig, Tim D{\"u}wel, Santanu Pal, Kalliopi Meladaki, Mahsa Monshizadeh, Antonio Kr{\"u}ger, and Josef van Genabith. 2020.
\newblock \href {https://doi.org/10.18653/v1/2020.acl-main.155} {{MMPE}: {A} {M}ulti-{M}odal {I}nterface for {P}ost-{E}diting {M}achine {T}ranslation}.
\newblock In \emph{Proceedings of the 58th Annual Meeting of the Association for Computational Linguistics}, pages 1691--1702. Association for Computational Linguistics.

\bibitem[{Himmi et~al.(2024)Himmi, Staerman, Picot, Colombo, and Guerreiro}]{himmi-etal-2024-enhanced}
Anas Himmi, Guillaume Staerman, Marine Picot, Pierre Colombo, and Nuno~M Guerreiro. 2024.
\newblock \href {https://doi.org/10.18653/v1/2024.emnlp-main.1033} {Enhanced hallucination detection in neural machine translation through simple detector aggregation}.
\newblock In \emph{Proceedings of the 2024 Conference on Empirical Methods in Natural Language Processing}, pages 18573--18583. Association for Computational Linguistics.

\bibitem[{Kepler et~al.(2019)Kepler, Tr{\'e}nous, Treviso, Vera, and Martins}]{kepler-etal-2019-openkiwi}
Fabio Kepler, Jonay Tr{\'e}nous, Marcos Treviso, Miguel Vera, and Andr{\'e} F.~T. Martins. 2019.
\newblock \href {https://doi.org/10.18653/v1/P19-3020} {{O}pen{K}iwi: An open source framework for quality estimation}.
\newblock In \emph{Proceedings of the 57th Annual Meeting of the Association for Computational Linguistics: System Demonstrations}, pages 117--122. Association for Computational Linguistics.

\bibitem[{Kocmi et~al.(2024{\natexlab{a}})Kocmi, Avramidis, Bawden, Bojar, Dvorkovich, Federmann, Fishel, Freitag, Gowda, Grundkiewicz, Haddow, Karpinska, Koehn, Marie, Monz, Murray, Nagata, Popel, Popovi{\'c}, Shmatova, Steingr{\'i}msson, and Zouhar}]{kocmi-etal-2024-findings}
Tom Kocmi, Eleftherios Avramidis, Rachel Bawden, Ond{\v{r}}ej Bojar, Anton Dvorkovich, Christian Federmann, Mark Fishel, Markus Freitag, Thamme Gowda, Roman Grundkiewicz, Barry Haddow, Marzena Karpinska, Philipp Koehn, Benjamin Marie, Christof Monz, Kenton Murray, Masaaki Nagata, Martin Popel, Maja Popovi{\'c}, Mariya Shmatova, Steinth{\'o}r Steingr{\'i}msson, and Vil{\'e}m Zouhar. 2024{\natexlab{a}}.
\newblock \href {https://doi.org/10.18653/v1/2024.wmt-1.1} {Findings of the {WMT}24 general machine translation shared task: The {LLM} era is here but {MT} is not solved yet}.
\newblock In \emph{Proceedings of the Ninth Conference on Machine Translation}, pages 1--46. Association for Computational Linguistics.

\bibitem[{Kocmi et~al.(2023)Kocmi, Avramidis, Bawden, Bojar, Dvorkovich, Federmann, Fishel, Freitag, Gowda, Grundkiewicz, Haddow, Koehn, Marie, Monz, Morishita, Murray, Nagata, Nakazawa, Popel, Popovi{\'c}, and Shmatova}]{kocmi-etal-2023-findings}
Tom Kocmi, Eleftherios Avramidis, Rachel Bawden, Ond{\v{r}}ej Bojar, Anton Dvorkovich, Christian Federmann, Mark Fishel, Markus Freitag, Thamme Gowda, Roman Grundkiewicz, Barry Haddow, Philipp Koehn, Benjamin Marie, Christof Monz, Makoto Morishita, Kenton Murray, Makoto Nagata, Toshiaki Nakazawa, Martin Popel, Maja Popovi{\'c}, and Mariya Shmatova. 2023.
\newblock \href {https://doi.org/10.18653/v1/2023.wmt-1.1} {Findings of the 2023 conference on machine translation ({WMT}23): {LLM}s are here but not quite there yet}.
\newblock In \emph{Proceedings of the Eighth Conference on Machine Translation}, pages 1--42. Association for Computational Linguistics.

\bibitem[{Kocmi and Federmann(2023)}]{kocmi-federmann-2023-gemba}
Tom Kocmi and Christian Federmann. 2023.
\newblock \href {https://doi.org/10.18653/v1/2023.wmt-1.64} {{GEMBA}-{MQM}: Detecting translation quality error spans with {GPT}-4}.
\newblock In \emph{Proceedings of the Eighth Conference on Machine Translation}, pages 768--775. Association for Computational Linguistics.

\bibitem[{Kocmi et~al.(2024{\natexlab{b}})Kocmi, Zouhar, Avramidis, Grundkiewicz, Karpinska, Popovi{\'c}, Sachan, and Shmatova}]{kocmi-etal-2024-error}
Tom Kocmi, Vil{\'e}m Zouhar, Eleftherios Avramidis, Roman Grundkiewicz, Marzena Karpinska, Maja Popovi{\'c}, Mrinmaya Sachan, and Mariya Shmatova. 2024{\natexlab{b}}.
\newblock \href {https://doi.org/10.18653/v1/2024.wmt-1.131} {Error span annotation: A balanced approach for human evaluation of machine translation}.
\newblock In \emph{Proceedings of the Ninth Conference on Machine Translation}, pages 1440--1453. Association for Computational Linguistics.

\bibitem[{Krings(2001)}]{krings2001repairing}
Hans~P Krings. 2001.
\newblock \href {https://www.erudit.org/fr/revues/meta/2002-v47-n3-meta693/008026ar/} {\emph{Repairing texts: Empirical investigations of machine translation post-editing processes}}.
\newblock Kent State University Press, Kent, Ohio and London.

\bibitem[{Leiter et~al.(2024)Leiter, Lertvittayakumjorn, Fomicheva, Zhao, Gao, and Eger}]{leiter-etal-2024-explainable}
Christoph Leiter, Piyawat Lertvittayakumjorn, Marina Fomicheva, Wei Zhao, Yang Gao, and Steffen Eger. 2024.
\newblock \href {http://jmlr.org/papers/v25/22-0416.html} {Towards explainable evaluation metrics for machine translation}.
\newblock \emph{Journal of Machine Learning Research}, 25(75):1--49.

\bibitem[{Li et~al.(2025)Li, Shi, Shang, Han, Zhao, Wang, Qian, Qian, Xu, Wu, Lyu, Wang, Tang, Luo, Xu, and Zhang}]{li-etal-2025-transbench}
Haijun Li, Tianqi Shi, Zifu Shang, Yuxuan Han, Xueyu Zhao, Hao Wang, Yu~Qian, Zhiqiang Qian, Linlong Xu, Minghao Wu, Chenyang Lyu, Longyue Wang, Gongbo Tang, Weihua Luo, Zhao Xu, and Kaifu Zhang. 2025.
\newblock \href {https://arxiv.org/abs/2505.14244} {Transbench: Benchmarking machine translation for industrial-scale applications}.
\newblock \emph{CoRR}, cs.CL/2505.14244v1.

\bibitem[{Lim et~al.(2024)Lim, Vylomova, Kemp, and Cohn}]{lim-etal-2024-predicting}
Zheng~Wei Lim, Ekaterina Vylomova, Charles Kemp, and Trevor Cohn. 2024.
\newblock \href {https://doi.org/10.1162/tacl\_a\_00714} {Predicting human translation difficulty with neural machine translation}.
\newblock \emph{Transactions of the Association for Computational Linguistics}, 12:1479--1496.

\bibitem[{Liu et~al.(2024)Liu, Riley, Deutsch, Lui, Niu, Shah, and Freitag}]{liu-etal-2024-beyond-human}
Zhongtao Liu, Parker Riley, Daniel Deutsch, Alison Lui, Mengmeng Niu, Apurva Shah, and Markus Freitag. 2024.
\newblock \href {https://doi.org/10.18653/v1/2024.wmt-1.110} {Beyond human-only: Evaluating human-machine collaboration for collecting high-quality translation data}.
\newblock In \emph{Proceedings of the Ninth Conference on Machine Translation}, pages 1095--1106. Association for Computational Linguistics.

\bibitem[{Lommel et~al.(2013)Lommel, Burchardt, and Uszkoreit}]{lommel-2013-multidimensional}
Arle~Richard Lommel, Aljoscha Burchardt, and Hans Uszkoreit. 2013.
\newblock \href {https://aclanthology.org/2013.tc-1.6/} {Multidimensional quality metrics: a flexible system for assessing translation quality}.
\newblock In \emph{Proceedings of Translating and the Computer 35}, London, UK. Aslib.

\bibitem[{Läubli et~al.(2022)Läubli, Simianer, Wuebker, Kovacs, Sennrich, and Green}]{laubli-etal-2021-impact}
Samuel Läubli, Patrick Simianer, Joern Wuebker, Geza Kovacs, Rico Sennrich, and Spence Green. 2022.
\newblock \href {https://doi.org/https://doi.org/10.1075/target.20006.lau} {The impact of text presentation on translator performance}.
\newblock \emph{Target. International Journal of Translation Studies}, 34(2):309--342.

\bibitem[{Mehandru et~al.(2023)Mehandru, Agrawal, Xiao, Gao, Khoong, Carpuat, and Salehi}]{mehandru-etal-2023-physician}
Nikita Mehandru, Sweta Agrawal, Yimin Xiao, Ge~Gao, Elaine Khoong, Marine Carpuat, and Niloufar Salehi. 2023.
\newblock \href {https://doi.org/10.18653/v1/2023.emnlp-main.712} {Physician detection of clinical harm in machine translation: Quality estimation aids in reliance and backtranslation identifies critical errors}.
\newblock In \emph{Proceedings of the 2023 Conference on Empirical Methods in Natural Language Processing}, pages 11633--11647. Association for Computational Linguistics.

\bibitem[{Moslem et~al.(2023)Moslem, Haque, Kelleher, and Way}]{moslem-etal-2023-adaptive}
Yasmin Moslem, Rejwanul Haque, John~D. Kelleher, and Andy Way. 2023.
\newblock \href {https://aclanthology.org/2023.eamt-1.22/} {Adaptive machine translation with large language models}.
\newblock In \emph{Proceedings of the 24th Annual Conference of the European Association for Machine Translation}, pages 227--237. European Association for Machine Translation.

\bibitem[{Neves et~al.(2024)Neves, Grozea, Thomas, Roller, Bawden, N{\'e}v{\'e}ol, Castle, Bonato, Di~Nunzio, Vezzani, Vicente~Navarro, Yeganova, and Jimeno~Yepes}]{neves-etal-2024-findings}
Mariana Neves, Cristian Grozea, Philippe Thomas, Roland Roller, Rachel Bawden, Aur{\'e}lie N{\'e}v{\'e}ol, Steffen Castle, Vanessa Bonato, Giorgio~Maria Di~Nunzio, Federica Vezzani, Maika Vicente~Navarro, Lana Yeganova, and Antonio Jimeno~Yepes. 2024.
\newblock \href {https://doi.org/10.18653/v1/2024.wmt-1.6} {Findings of the {WMT} 2024 biomedical translation shared task: Test sets on abstract level}.
\newblock In \emph{Proceedings of the Ninth Conference on Machine Translation}, pages 124--138. Association for Computational Linguistics.

\bibitem[{Neves et~al.(2023)Neves, Jimeno~Yepes, N{\'e}v{\'e}ol, Bawden, Di~Nunzio, Roller, Thomas, Vezzani, Vicente~Navarro, Yeganova, Wiemann, and Grozea}]{neves-etal-2023-findings}
Mariana Neves, Antonio Jimeno~Yepes, Aur{\'e}lie N{\'e}v{\'e}ol, Rachel Bawden, Giorgio~Maria Di~Nunzio, Roland Roller, Philippe Thomas, Federica Vezzani, Maika Vicente~Navarro, Lana Yeganova, Dina Wiemann, and Cristian Grozea. 2023.
\newblock \href {https://doi.org/10.18653/v1/2023.wmt-1.2} {Findings of the {WMT} 2023 biomedical translation shared task: Evaluation of {C}hat{GPT} 3.5 as a comparison system}.
\newblock In \emph{Proceedings of the Eighth Conference on Machine Translation}, pages 43--54. Association for Computational Linguistics.

\bibitem[{{NLLB Team} et~al.(2024){NLLB Team}, Costa-jussà, Cross, Çelebi, Elbayad, Heafield, Heffernan, Kalbassi, Lam, Licht, Maillard, Sun, Wang, Wenzek, Youngblood, Akula, Barrault, Gonzalez, Hansanti, Hoffman, Jarrett, Sadagopan, Rowe, Spruit, Tran, Andrews, Ayan, Bhosale, Edunov, Fan, Gao, Goswami, Guzmán, Koehn, Mourachko, Ropers, Saleem, Schwenk, and Wang}]{nllb}
{NLLB Team}, Marta~R. Costa-jussà, James Cross, Onur Çelebi, Maha Elbayad, Kenneth Heafield, Kevin Heffernan, Elahe Kalbassi, Janice Lam, Daniel Licht, Jean Maillard, Anna Sun, Skyler Wang, Guillaume Wenzek, Al~Youngblood, Bapi Akula, Loic Barrault, Gabriel~Mejia Gonzalez, Prangthip Hansanti, John Hoffman, Semarley Jarrett, Kaushik~Ram Sadagopan, Dirk Rowe, Shannon Spruit, Chau Tran, Pierre Andrews, Necip~Fazil Ayan, Shruti Bhosale, Sergey Edunov, Angela Fan, Cynthia Gao, Vedanuj Goswami, Francisco Guzmán, Philipp Koehn, Alexandre Mourachko, Christophe Ropers, Safiyyah Saleem, Holger Schwenk, and Jeff Wang. 2024.
\newblock \href {https://doi.org/10.1038/s41586-024-07335-x} {Scaling neural machine translation to 200 languages}.
\newblock \emph{Nature}, 630(8018):841--846.

\bibitem[{Rei et~al.(2022)Rei, C.~de Souza, Alves, Zerva, Farinha, Glushkova, Lavie, Coheur, and Martins}]{rei-etal-2022-comet}
Ricardo Rei, Jos{\'e}~G. C.~de Souza, Duarte Alves, Chrysoula Zerva, Ana~C Farinha, Taisiya Glushkova, Alon Lavie, Luisa Coheur, and Andr{\'e} F.~T. Martins. 2022.
\newblock \href {https://aclanthology.org/2022.wmt-1.52/} {{COMET}-22: Unbabel-{IST} 2022 submission for the metrics shared task}.
\newblock In \emph{Proceedings of the Seventh Conference on Machine Translation (WMT)}, pages 578--585. Association for Computational Linguistics.

\bibitem[{Rei et~al.(2021)Rei, Farinha, Zerva, van Stigt, Stewart, Ramos, Glushkova, Martins, and Lavie}]{rei-etal-2021-references}
Ricardo Rei, Ana~C Farinha, Chrysoula Zerva, Daan van Stigt, Craig Stewart, Pedro Ramos, Taisiya Glushkova, Andr{\'e} F.~T. Martins, and Alon Lavie. 2021.
\newblock \href {https://aclanthology.org/2021.wmt-1.111/} {Are references really needed? unbabel-{IST} 2021 submission for the metrics shared task}.
\newblock In \emph{Proceedings of the Sixth Conference on Machine Translation}, pages 1030--1040. Association for Computational Linguistics.

\bibitem[{Rei et~al.(2020)Rei, Stewart, Farinha, and Lavie}]{rei-etal-2020-comet}
Ricardo Rei, Craig Stewart, Ana~C Farinha, and Alon Lavie. 2020.
\newblock \href {https://doi.org/10.18653/v1/2020.emnlp-main.213} {{COMET}: A neural framework for {MT} evaluation}.
\newblock In \emph{Proceedings of the 2020 Conference on Empirical Methods in Natural Language Processing (EMNLP)}, pages 2685--2702. Association for Computational Linguistics.

\bibitem[{Sarti et~al.(2022)Sarti, Bisazza, Guerberof-Arenas, and Toral}]{sarti-etal-2022-divemt}
Gabriele Sarti, Arianna Bisazza, Ana Guerberof-Arenas, and Antonio Toral. 2022.
\newblock \href {https://doi.org/10.18653/v1/2022.emnlp-main.532} {{D}iv{EMT}: Neural machine translation post-editing effort across typologically diverse languages}.
\newblock In \emph{Proceedings of the 2022 Conference on Empirical Methods in Natural Language Processing}, pages 7795--7816. Association for Computational Linguistics.

\bibitem[{Savoldi et~al.(2025)Savoldi, Ramponi, Negri, and Bentivogli}]{savoldi-etal-2025-translation}
Beatrice Savoldi, Alan Ramponi, Matteo Negri, and Luisa Bentivogli. 2025.
\newblock \href {https://arxiv.org/abs/2502.13780} {Translation in the hands of many: Centering lay users in machine translation interactions}.
\newblock \emph{CoRR}, cs.CL/2502.13780v1.

\bibitem[{Shenoy et~al.(2021)Shenoy, Herbig, Kr{\"u}ger, and van Genabith}]{shenoy-etal-2021-investigating}
Raksha Shenoy, Nico Herbig, Antonio Kr{\"u}ger, and Josef van Genabith. 2021.
\newblock \href {https://doi.org/10.18653/v1/2021.emnlp-main.799} {Investigating the helpfulness of word-level quality estimation for post-editing machine translation output}.
\newblock In \emph{Proceedings of the 2021 Conference on Empirical Methods in Natural Language Processing}, pages 10173--10185. Association for Computational Linguistics.

\bibitem[{Specia et~al.(2020)Specia, Blain, Fomicheva, Fonseca, Chaudhary, Guzm{\'a}n, and Martins}]{specia-etal-2020-findings-wmt}
Lucia Specia, Fr{\'e}d{\'e}ric Blain, Marina Fomicheva, Erick Fonseca, Vishrav Chaudhary, Francisco Guzm{\'a}n, and Andr{\'e} F.~T. Martins. 2020.
\newblock \href {https://aclanthology.org/2020.wmt-1.79/} {Findings of the {WMT} 2020 shared task on quality estimation}.
\newblock In \emph{Proceedings of the Fifth Conference on Machine Translation}, pages 743--764. Association for Computational Linguistics.

\bibitem[{Specia et~al.(2009)Specia, Turchi, Cancedda, Cristianini, and Dymetman}]{specia-etal-2009-estimating}
Lucia Specia, Marco Turchi, Nicola Cancedda, Nello Cristianini, and Marc Dymetman. 2009.
\newblock \href {https://aclanthology.org/2009.eamt-1.5/} {Estimating the sentence-level quality of machine translation systems}.
\newblock In \emph{Proceedings of the 13th Annual Conference of the European Association for Machine Translation}. European Association for Machine Translation.

\bibitem[{Sun et~al.(2022)Sun, Swayamdipta, May, and Ma}]{sun-etal-2022-investigating}
Jiao Sun, Swabha Swayamdipta, Jonathan May, and Xuezhe Ma. 2022.
\newblock \href {https://doi.org/10.18653/v1/2022.findings-emnlp.432} {Investigating the benefits of free-form rationales}.
\newblock In \emph{Findings of the Association for Computational Linguistics: EMNLP 2022}, pages 5867--5882. Association for Computational Linguistics.

\bibitem[{Tamchyna(2021)}]{tamchyna-2021-deploying}
Ale{\v{s}} Tamchyna. 2021.
\newblock \href {https://aclanthology.org/2021.mtsummit-up.21/} {Deploying {MT} quality estimation on a large scale: Lessons learned and open questions}.
\newblock In \emph{Proceedings of Machine Translation Summit XVIII: Users and Providers Track}, pages 291--305. Association for Machine Translation in the Americas.

\bibitem[{Tang et~al.(2021)Tang, Tran, Li, Chen, Goyal, Chaudhary, Gu, and Fan}]{tang-etal-2021-multilingual}
Yuqing Tang, Chau Tran, Xian Li, Peng-Jen Chen, Naman Goyal, Vishrav Chaudhary, Jiatao Gu, and Angela Fan. 2021.
\newblock \href {https://doi.org/10.18653/v1/2021.findings-acl.304} {Multilingual translation from denoising pre-training}.
\newblock In \emph{Findings of the Association for Computational Linguistics: ACL-IJCNLP 2021}, pages 3450--3466. Association for Computational Linguistics.

\bibitem[{Thompson and Post(2020)}]{thompson-post-2020-automatic}
Brian Thompson and Matt Post. 2020.
\newblock \href {https://doi.org/10.18653/v1/2020.emnlp-main.8} {Automatic machine translation evaluation in many languages via zero-shot paraphrasing}.
\newblock In \emph{Proceedings of the 2020 Conference on Empirical Methods in Natural Language Processing (EMNLP)}, pages 90--121. Association for Computational Linguistics.

\bibitem[{Turchi et~al.(2014)Turchi, Anastasopoulos, C.~de Souza, and Negri}]{turchi-etal-2014-adaptive}
Marco Turchi, Antonios Anastasopoulos, Jos{\'e}~G. C.~de Souza, and Matteo Negri. 2014.
\newblock \href {https://doi.org/10.3115/v1/P14-1067} {Adaptive quality estimation for machine translation}.
\newblock In \emph{Proceedings of the 52nd Annual Meeting of the Association for Computational Linguistics (Volume 1: Long Papers)}, pages 710--720. Association for Computational Linguistics.

\bibitem[{Turchi et~al.(2013)Turchi, Negri, and Federico}]{turchi-etal-2013-coping}
Marco Turchi, Matteo Negri, and Marcello Federico. 2013.
\newblock \href {https://aclanthology.org/W13-2231/} {Coping with the subjectivity of human judgements in {MT} quality estimation}.
\newblock In \emph{Proceedings of the Eighth Workshop on Statistical Machine Translation}, pages 240--251. Association for Computational Linguistics.

\bibitem[{Vasconcelos et~al.(2025)Vasconcelos, Bansal, Fourney, Liao, and Wortman~Vaughan}]{vasconcelos-etal-2024-generation}
Helena Vasconcelos, Gagan Bansal, Adam Fourney, Q.~Vera Liao, and Jennifer Wortman~Vaughan. 2025.
\newblock \href {https://doi.org/10.1145/3702320} {Generation probabilities are not enough: Uncertainty highlighting in ai code completions}.
\newblock \emph{ACM Trans. Comput.-Hum. Interact.}, 32(1).

\bibitem[{Xu et~al.(2023)Xu, Agrawal, Briakou, Martindale, and Carpuat}]{xu-etal-2023-understanding}
Weijia Xu, Sweta Agrawal, Eleftheria Briakou, Marianna~J. Martindale, and Marine Carpuat. 2023.
\newblock \href {https://doi.org/10.1162/tacl\_a\_00563} {Understanding and detecting hallucinations in neural machine translation via model introspection}.
\newblock \emph{Transactions of the Association for Computational Linguistics}, 11:546--564.

\bibitem[{Zerva et~al.(2024)Zerva, Blain, C.~De~Souza, Kanojia, Deoghare, Guerreiro, Attanasio, Rei, Orasan, Negri, Turchi, Chatterjee, Bhattacharyya, Freitag, and Martins}]{zerva-etal-2024-findings}
Chrysoula Zerva, Frederic Blain, Jos{\'e}~G. C.~De~Souza, Diptesh Kanojia, Sourabh Deoghare, Nuno~M. Guerreiro, Giuseppe Attanasio, Ricardo Rei, Constantin Orasan, Matteo Negri, Marco Turchi, Rajen Chatterjee, Pushpak Bhattacharyya, Markus Freitag, and Andr{\'e} Martins. 2024.
\newblock \href {https://doi.org/10.18653/v1/2024.wmt-1.3} {Findings of the quality estimation shared task at {WMT} 2024: Are {LLM}s closing the gap in {QE}?}
\newblock In \emph{Proceedings of the Ninth Conference on Machine Translation}, pages 82--109. Association for Computational Linguistics.

\bibitem[{Zerva et~al.(2022)Zerva, Blain, Rei, Lertvittayakumjorn, C.~de Souza, Eger, Kanojia, Alves, Or{\u{a}}san, Fomicheva, Martins, and Specia}]{zerva-etal-2022-findings}
Chrysoula Zerva, Fr{\'e}d{\'e}ric Blain, Ricardo Rei, Piyawat Lertvittayakumjorn, Jos{\'e}~G. C.~de Souza, Steffen Eger, Diptesh Kanojia, Duarte Alves, Constantin Or{\u{a}}san, Marina Fomicheva, Andr{\'e} F.~T. Martins, and Lucia Specia. 2022.
\newblock \href {https://aclanthology.org/2022.wmt-1.3/} {Findings of the {WMT} 2022 shared task on quality estimation}.
\newblock In \emph{Proceedings of the Seventh Conference on Machine Translation (WMT)}, pages 69--99. Association for Computational Linguistics.

\bibitem[{Zouhar et~al.(2024)Zouhar, Ding, Currey, Badeka, Wang, and Thompson}]{zouhar-etal-2024-fine}
Vil{\'e}m Zouhar, Shuoyang Ding, Anna Currey, Tatyana Badeka, Jenyuan Wang, and Brian Thompson. 2024.
\newblock \href {https://doi.org/10.18653/v1/2024.acl-short.45} {Fine-tuned machine translation metrics struggle in unseen domains}.
\newblock In \emph{Proceedings of the 62nd Annual Meeting of the Association for Computational Linguistics (Volume 2: Short Papers)}, pages 488--500. Association for Computational Linguistics.

\bibitem[{Zouhar et~al.(2025)Zouhar, Kocmi, and Sachan}]{zouhar-etal-2025-ai}
Vil{\'e}m Zouhar, Tom Kocmi, and Mrinmaya Sachan. 2025.
\newblock \href {https://doi.org/10.18653/v1/2025.naacl-long.255} {{AI}-assisted human evaluation of machine translation}.
\newblock In \emph{Proceedings of the 2025 Conference of the Nations of the Americas Chapter of the Association for Computational Linguistics: Human Language Technologies (Volume 1: Long Papers)}, pages 4936--4950, Albuquerque, New Mexico. Association for Computational Linguistics.

\bibitem[{Zouhar et~al.(2021{\natexlab{a}})Zouhar, Nov{\'a}k, {\v{Z}}ilinec, Bojar, Obreg{\'o}n, Hill, Blain, Fomicheva, Specia, and Yankovskaya}]{zouhar-etal-2021-backtranslation}
Vil{\'e}m Zouhar, Michal Nov{\'a}k, Mat{\'u}{\v{s}} {\v{Z}}ilinec, Ond{\v{r}}ej Bojar, Mateo Obreg{\'o}n, Robin~L. Hill, Fr{\'e}d{\'e}ric Blain, Marina Fomicheva, Lucia Specia, and Lisa Yankovskaya. 2021{\natexlab{a}}.
\newblock \href {https://doi.org/10.18653/v1/2021.naacl-main.14} {Backtranslation feedback improves user confidence in {MT}, not quality}.
\newblock In \emph{Proceedings of the 2021 Conference of the North American Chapter of the Association for Computational Linguistics: Human Language Technologies}, pages 151--161. Association for Computational Linguistics.

\bibitem[{Zouhar et~al.(2021{\natexlab{b}})Zouhar, Popel, Bojar, and Tamchyna}]{zouhar-etal-2021-neural}
Vil{\'e}m Zouhar, Martin Popel, Ond{\v{r}}ej Bojar, and Ale{\v{s}} Tamchyna. 2021{\natexlab{b}}.
\newblock \href {https://doi.org/10.18653/v1/2021.emnlp-main.801} {Neural machine translation quality and post-editing performance}.
\newblock In \emph{Proceedings of the 2021 Conference on Empirical Methods in Natural Language Processing}, pages 10204--10214. Association for Computational Linguistics.

\end{thebibliography}
\bibliographystyle{misc/acl_natbib}

\clearpage
\appendix

\begin{figure*}
    \centering

    {
    \tiny
    \begin{tabularx}{\linewidth}{p{0.7cm}p{2.45cm}p{2.6cm}p{2.8cm}|p{1.6cm}cc|c}
        \toprule[1.5pt]
                \textbf{Doc ID - Seg. ID} & \multirow{2}{*}{\textbf{Source text}} & \multirow{2}{*}{\textbf{Target text}} & \multirow{2}{*}{\textbf{Proposed correction}} & \multicolumn{3}{c}{\textbf{Error Annotation}} & \multirow{2}{*}{\textbf{Score}} \\
                \cmidrule(lr){5-7}
                & & & & \textbf{Description} & \textbf{Category} & \textbf{Severity} & \\
        \midrule
        9-1 & Specifying peri- and postnatal factors in children born very preterm (VPT) that affect later outcome helps to improve long-term treatment. & Specificare i fattori peri- e postnatali nei bambini nati molto pretermine (\textcolor{red}{VPT}) che influenzano il risultato successivo aiuta a migliorare il trattamento a lungo termine. & Specificare i fattori peri- e postnatali nei bambini nati molto pretermine (\textcolor{red}{VPT, Very Preterm}) che influenzano il risultato successivo aiuta a migliorare il trattamento a lungo termine. & When we have a foreign acronym, the usual rule is to indicate also the whole term the first time it appears. & Readability & Minor & 90 \\
        \midrule
        9-2 & To enhance the predictability of 5-year cognitive outcome by perinatal, 2-year developmental and socio-economic data. & Migliorare la prevedibilità del risultato cognitivo a 5 anni mediante dati perinatali, di sviluppo e socioeconomici a 2 anni. & & & & & 100 \\
        \midrule
        9-3 & 5-year infants born VPT were compared to 34 term controls.
        & 
        I neonati di 5 anni nati VPT sono stati confrontati con 34 \textcolor{red}{nati a termine come controllo}.
        & 
        I neonati di 5 anni nati VPT sono stati confrontati con 34 \textcolor{red}{controlli a termine}. & & Mistranslation & Minor & 70 \\
        \midrule
        \multirow{2}{*}{9-4} & \multirow{2}{2.45cm}{The IQ of 5-year infants born VPT was 10 points lower than that of term controls and influenced independently by preterm birth and SES.} & Il QI dei bambini di 5 anni nati VPT era di 10 punti inferiore a quello dei nati a termine \textcolor{red}{di controllo,} e influenzato indipendentemente dalla nascita pretermine e dai dati SES. & Il QI dei bambini di 5 anni nati VPT era di 10 punti inferiore a quello dei nati a termine e influenzato indipendentemente dalla nascita pretermine e dallo stato socioeconomico (SES). & & Mistranslation & Minor & 70 \\
        \cmidrule(lr){3-7}
         & & Il QI dei bambini di 5 anni nati VPT era di 10 punti inferiore a quello dei nati a termine di controllo, e influenzato indipendentemente dalla nascita pretermine e \textcolor{red}{dai dati SES}. & Il QI dei bambini di 5 anni nati VPT era di 10 punti inferiore a quello dei nati a termine e influenzato indipendentemente dalla nascita pretermine e \textcolor{red}{dallo stato socioeconomico (SES)}. & Unexplained acronym. Non-expert people could have trouble understanding the meaning. & Untranslated & Minor &  \\
        \midrule
        \midrule
        52-1 & But with less than 3 months to go for that, I feel I’m not ready yet, but having never taken it, I have nothing to compare it to besides colleagues’ advice. 
        & Ma con meno di 3 mesi per farlo, sento di non essere ancora pronto, \textcolor{red}{ma non l'ho mai preso}, non ho nulla con cui confrontarlo oltre ai consigli dei colleghi. 
        & Ma con meno di 3 mesi per farlo, sento di non essere ancora pronto, \textcolor{red}{e non avendolo mai fatto}, non ho nulla con cui confrontarlo oltre ai consigli dei colleghi.& & Mistranslation & Major & 30 \\
        \midrule
        \multirow{2}{*}{52-2} & \multirow{2}{2.45cm}{Without knowing what I know, they can’t know if I’m actually ready yet, but many of them are pushing me to sign up for it.}
        & \textcolor{red}{Senza sapere quello che so}, non possono sapere se sono ancora pronta, ma molti di loro mi stanno spingendo a iscrivermi.
        & \textcolor{red}{Se non hanno idea di quanto sappia}, non possono sapere se sono davvero pronta, ma molti di loro mi stanno spingendo a iscrivermi. & & Readability & Minor & 60 \\
        \cmidrule(lr){3-7}
        & & Senza sapere quello che so, non possono sapere se sono \textcolor{red}{ancora} pronta, ma molti di loro mi stanno spingendo a iscrivermi.
        & Se non hanno idea di quanto sappia, non possono sapere se sono \textcolor{red}{davvero} pronta, ma molti di loro mi stanno spingendo a iscrivermi. & & Mistranslation & Minor & \\
        \midrule
        52-3 & I’m close... I just don’t know if I’m 2 months study close. & Ci sono quasi... solo che non so se ce la farò in soli 2 mesi\textcolor{red}{, ma penso di potercela fare}. & Ci sono quasi... solo che non so se ce la farò in soli 2 mesi. & & Addition & Major & 20 \\
        \bottomrule[1.5pt]
    \end{tabularx}
    }
    {
    \tiny
    \begin{tabularx}{\linewidth}{llX}
    \bf Error category &
    \bf Subcategory &
    \bf Description \\
    \midrule
    \multirow{5}{*}{\makecell*[{{p{3cm}}}]{\textbf{Accuracy}\newline Incorrect meaning has been transferred to the source text.}}
    & \textbf{Addition}
    & Translation includes the information that is not present in the source and it changes or distorts the original message. \\
    \cmidrule(lr){3-3}
    & \textbf{Omission} & Translation is missing the information that is present in the source, which is important to convey the message. \\
    \cmidrule(lr){3-3}
    & \textbf{Mistranslation} & Translation does not accurately represent the source content meaning. \\
    \cmidrule(lr){3-3}
    & \textbf{Inconsistency} & There are internal inconsistencies in the translation (for example, using different verb forms in the bullet list or in CTAs, calling the same UI element differently, terminology used inconsistently etc). \\
    \cmidrule(lr){3-3}
    & \textbf{Untranslated} & Content that should have been translated has been left untranslated. \\
    \midrule
    \multirow{3}{*}{\makecell*[{{p{3cm}}}]{\textbf{Linguistic}\newline Official linguistic reference sources such as grammar books.}}
    & \textbf{Punctuation} & Punctuation is used incorrectly (for the locale or style), including missing or extra white spaces and the incorrect use of space (non-breaking space). Violation of typographic conventions of the locale. \\
    \cmidrule(lr){3-3}
    & \textbf{Spelling} & Issues related to spelling of words, including typos, wrong word hyphenation, word breaks and capitalization. \\
    \cmidrule(lr){3-3}
    & \textbf{Grammar} & Issues related to the grammar or syntax of the text, other than spelling. \\
    \midrule
    \multirow{3}{*}{\makecell*[{{p{3cm}}}]{\textbf{Style}\newline Not suitable/native; too literal or awkward.}}
    & \textbf{Inconsistent Style} & Style is inconsistent within a text. \\
    \cmidrule(lr){3-3}
    & \textbf{Readability} & Translation does not read well (due to heavy sentence structure, frequent repetitions, unidiomatic). \\
    \cmidrule(lr){3-3}
    & \textbf{Wrong Register} & Inappropriate style for the specific subject field, the level of formality, and the mode of discourse (e.g., written text versus transcribed speech). \\
    \end{tabularx}
    \begin{tabularx}{\linewidth}{lX}
    \toprule
    \textbf{Severity level} & \textbf{Description}\\
    \midrule
    \textbf{Major} & The Severity Level of an error that seriously affects the understandability, reliability, or usability of the content for its intended purpose or hinders the proper use of the product or service due to a significant loss or change in meaning or because the error appears in a highly visible or important part of the content.\\
    \cmidrule(lr){2-2}
    \textbf{Minor} & The Severity Level of an error that does not seriously impede the usability, understandability, or reliability of the content for its intended purpose, but has a limited impact on, for example, accuracy, stylistic quality, consistency, fluency, clarity, or general appeal of the content.\\
    \cmidrule(lr){2-2}
    \textbf{Neutral} & The Severity Level of an error that differs from a quality evaluator's preferential translation or that is flagged for the translator’s attention but is an acceptable translation.\\
    \bottomrule
    \end{tabularx}
    }
    
    \caption{\textbf{Top:}QA interface with cropped examples of biomedical and social media texts with error annotations (Biomedical: post-edited segments with \NoHighlight; Social media: MT outputs). \textbf{Bottom:} Annotation instructions for our MQM-inspired error taxonomy.}
    \label{fig:qa}
\end{figure*}

\section{Filtering Details for QE4PE Data}
\label{app:data_stats}

\begin{enumerate}[noitemsep,left=0mm,topsep=0mm]
\item \textit{Documents should contain between 4 and 10 segments, each containing 10-100 words (959 docs).} This ensures that all documents are roughly uniform in terms of size and complexity to maintain a steady editing flow (\Cref{sec:interface}).
\item \textit{The average segment-level QE score predicted by XCOMET-XXL is between 0.3 and 0.95, with no segment below 0.3 (429 docs).} This forces segments to have a decent but still imperfect quality, excluding fully wrong translations.
\item \textit{At least 3 and at most 20 errors spans per document, with no more than 30\% of words in the document being highlighted (351 docs).} This avoids overwhelming the editor with excessive highlighting, while still ensuring error presence.
\end{enumerate}

The same heuristics were applied to both translation directions, selecting only documents matching our criteria in both cases.

\begin{table}[t]
    \footnotesize
    \centering
    \begin{tabular}{p{1cm}p{5.5cm}}
    \toprule[1.5pt]
    \multicolumn{2}{l}{\textbf{Remove negation (13-6)}} \\
    \midrule
    \textbf{English} & \textcolor{red}{No significant differences} were found with respect to principal diagnoses [...] \\
    \textbf{Dutch} & Er werden \textcolor{red}{geen significante verschillen $\rightarrow$ significante verschillen} gevonden met betrekking tot de belangrijkste diagnoses [...] \\
    \midrule
    \multicolumn{2}{l}{\textbf{Title literal translation (16-3)}} \\
    \midrule
    \textbf{English} & \textcolor{red}{The Last of Us} is an easy and canonical example of dad-ification. [...] \\
    \textbf{Italian} & \textcolor{red}{The Last of Us $\rightarrow$ L'ultimo di noi} è un esempio facile e canonico di dad-ification. [...] \\
    \midrule
    \multicolumn{2}{l}{\textbf{Wrong term (48-5)}} \\
    \midrule
    \textbf{English} & [...], , except for \textcolor{red}{alkaline phosphatase}. \\
    \textbf{Italian} & [...], ad eccezione della \textcolor{red}{fosfatasi alcalina $\rightarrow$ chinasi proteica}. \\
    \bottomrule[1.5pt]
    \end{tabular}
    \caption{Examples of original $\rightarrow$ manually inserted critical errors with document-segment ID from~\Cref{tab:critical-errors}.}
    \label{tab:critical-errors-examples}\vspace{-10pt}
\end{table}

\begin{figure}[t]
\includegraphics[width=\linewidth]{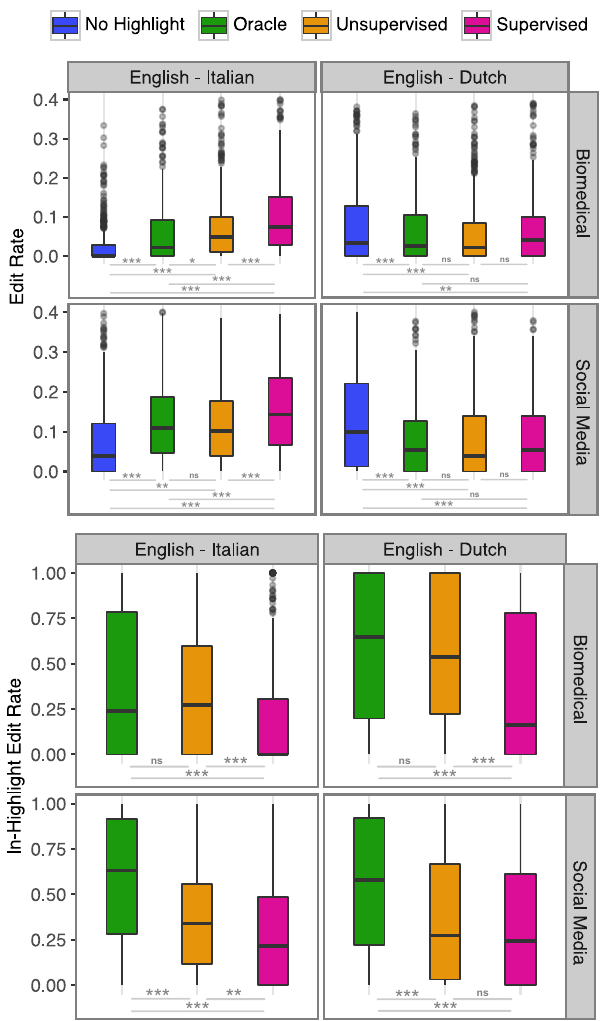}
\caption{\textbf{Top:} Post-editing rate across highlight modalities, domains and directions. \textbf{Bottom:} Proportion of edits in highlighted spans across highlight modalities. *** $=p<0.001$, ** $=p<0.01$, * $=p<0.05$, ns $=$ not significant with Bonferroni correction.}\vspace{-12pt}
\label{fig:editing}
\end{figure}

\begin{table}[t]
    \footnotesize
    \centering
    \begin{tabular}{lcc}
    \toprule[1.5pt]
    \multicolumn{3}{l}{\textbf{Target:} Seg. Edit Time, 5s bins from 0 to 600s} \\[0.2em]
    \textbf{Feature} & \textbf{Coeff.} & \textbf{Significance} \\
    \midrule
    (Intercept)                       & 1.67 & *** \\
    MT Num. Chars                    & 2.42 & *** \\
    Highlight Ratio \%               & 1.59 & *** \\
    Target Lang.: ITA                & -0.34 & *** \\
    Text Domain: Social              & 0.31 & *** \\
    \OracleShort~Highlight           & -0.79 & . \\
    \SupervisedShort Highlight       & 0.02 & \\
    \UnsupervisedShort~Highlight     & -0.07 & \\
    MT XCOMET QE Score               & 0.01 & *** \\
    ITA:\OracleShort                 & 0.91 & *** \\
    ITA:\SupervisedShort             & 1.18 & *** \\
    ITA:\UnsupervisedShort           & 0.48 & *** \\
    Social:\OracleShort              & -0.19 & ** \\
    Social:\SupervisedShort          & -0.34 & *** \\
    Social:\UnsupervisedShort        & -0.22 & *** \\
    Highlight Ratio:\OracleShort     & -0.83 & * \\
    Highlight Ratio:\SupervisedShort & -1.33 & *** \\
    \midrule
    Edit Order & \multicolumn{2}{c}{\multirow{3}{*}{\textbf{Random Factors}}} \\
    Translator ID \\
    Segment ID \\
    \bottomrule[1.5pt]
    \end{tabular}
    \caption{Details for the negative binomial mixed-effect model used for the productivity analysis of \Cref{sec:productivity}.}
    \label{tab:edit_time_model}
\end{table}

\begin{table}[t]
    \footnotesize
    \centering
    \begin{tabular}{lcc}
    \toprule[1.5pt]
    \multicolumn{3}{l}{\textbf{Target:} \% of edited characters in a segment (0-100).} \\
    \midrule
    \textbf{Feature} & \textbf{Coeff.} & \textbf{Significance} \\
    \midrule
    (Intercept)                      & 21.0  & *** \\
    MT Num. Chars                    & 10.3  & *** \\
    Highlight Ratio \%               & 7.1   & *** \\
    Target Lang.: ITA                & -9.9  & *** \\
    Text Domain: Social              & 10.9  & *** \\
    \OracleShort~Highlight           & -5.2  & \\
    \SupervisedShort Highlight       & -4.7  & \\
    \UnsupervisedShort~Highlight     & -0.9  & \\
    ITA:\OracleShort                 & 12.2 & *** \\
    ITA:\SupervisedShort             & 15.9 & *** \\
    ITA:\UnsupervisedShort           & 13.4 & *** \\
    Social:\OracleShort              & 3.5  & *** \\
    Social:\SupervisedShort          & -0.4 &  \\
    Social:\UnsupervisedShort        & 2.1  & ** \\
    Highlight Ratio:\OracleShort     & -0.18 & \\
    Highlight Ratio:\SupervisedShort & -1.78 & *** \\
    \midrule
    Edit Order & \multicolumn{2}{c}{\multirow{3}{*}{\textbf{Random Factors}}} \\
    Translator ID \\
    Segment ID \\
    \midrule
    MT Num. Chars & \multicolumn{2}{c}{\multirow{4}{*}{\textbf{Zero-Inflation Factors}}} \\
    Target Lang \\
    Text Domain \\
    Translator ID \\
    \bottomrule[1.5pt]
    \end{tabular}
    \caption{Details for the zero-inflated negative binomial mixed-effect model used for the editing analysis of \Cref{sec:highlights-edits}. The model achieves an RMSE of 0.11 and an $R^2$ of 0.98.}
    \label{tab:edit_rate_model}
\end{table}

\begin{sidewaystable*}
    \centering
    \tiny
    \begin{tabularx}{\linewidth}{lllllllXX}
    \toprule[1.5pt]
        \textbf{Identifier} & \textbf{Job} & \textbf{Eng. Lvl} & \textbf{Trans. YoE} & \textbf{Post-edit YoE} & \textbf{Post-edit \%} & \textbf{Adv. CAT YoE} & \textbf{MT good/bad for:} & \textbf{Post-edit comment} \\
        \midrule
        \texttt{eng-ita-nohigh-fast} & Freelance (FT) & C1 & 2-5  & 2-5  & 100\% & Often     & Good: Productivity, quality, repetitive work. & PE better than from scratch when consistency is needed. \\
        \texttt{eng-ita-nohigh-avg}  & Freelance (PT) & C1 & >10  & <2   & 20\%  & Often     & Good: Productivity, repetitive work. Bad: less creative. & PE produces unnatural sentences. \\
        \texttt{eng-ita-nohigh-slow} & Freelance (PT) & C2 & >10  & 2-5  & 40\%  & Sometimes & Good: creativity. & Good for time saving. \\
        \texttt{eng-ita-oracle-fast} & Freelance (FT) & C2 & 5-10 & 2-5  & 60\%  & Sometimes & Good: Productivity, repetitive work. Bad: less creative. & Good for productivity, humans always needed. \\
        \texttt{eng-ita-oracle-avg}  & Freelance (FT) & C2 & 5-10 & 5-10 & 20\%  & Always    & Good: productivity, terminology. & Good for tech docs, not for articulated texts. \\
        \texttt{eng-ita-oracle-slow} & Freelance (FT) & C2 & 2-5  & 5-10 & 80\%  & Always    & Good: Productivity, repetitive work. & Useful for consistency and productivity, unless creativity is needed. \\
        \texttt{eng-ita-unsup-fast}  & Freelance (FT) & C1 & <2   & <2   & 60\%  & Often     & Good: Productivity, terminology. Bad: less creative. & Humans will always be needed in translation. \\
        \texttt{eng-ita-unsup-avg}   & Freelance (FT) & C1 & >10  & 2-5  & 60\%  & Often     & Good: Productivity, repetitive work. Bad: less creative. & An opportunity for translators. \\
        \texttt{eng-ita-unsup-slow}  & Freelance (FT) & C1 & 5-10 & 5-10 & 80\%  & Always    & Good: Productivity, repetitive work. Bad: less creative. & Good for focusing on detailed/cultural/creative aspects of translations. \\
        \texttt{eng-ita-sup-fast}    & Freelance (PT) & C1 & >10  & 2-5  & 40\%  & Often     & Good: Productivity, quality, repetitive work, terminology. & Improves quality and consistency. \\
        \texttt{eng-ita-sup-avg}     & Freelance (FT) & C1 & >10  & 5-10 & 100\% & Always    & Good: Productivity, repetitive work. Bad: less creative. & Consistency improved, but less variance means less creativity. \\
        \texttt{eng-ita-sup-slow}    & Freelance (FT) & C1 & >10  & 2-5  & 20\%  & Always    & Good: Productivity, creativity, quality, repetitive work. & Good for productivity, but does not work on creative texts. \\
    \midrule
        \texttt{eng-nld-nohigh-fast} & Freelance (FT) & C1 & >10 & >10 & 40\% & Often & Good: Productivity, terminology. Bad: creativity. & Widespread but still too literal \\
        \texttt{eng-nld-nohigh-avg}  & Freelance (FT) & C2 & >10 & 2-5 & 40\% & Always & Good: Repetitive work. Bad: creativity, often wrong, worse quality. & Increase in productivity to save on costs brings down quality. \\
        \texttt{eng-nld-nohigh-slow} & Freelance (FT) & C2 & >10 & 5-10 & 100\% & Often & Good: Creativity, quality, repetitive work, terminology. & Working with MT can be creative beyond PE. \\
        \texttt{eng-nld-oracle-fast} & Freelance (FT) & C1 & 5-10 & 5-10 & 80\% & Always & Good: Productivity, quality, repetitive work, terminology. & Good for tech docs and repetition. \\
        \texttt{eng-nld-oracle-avg}  & Freelance (FT) & C2 & >10 & 2-5 & 40\% & Always & Bad: less creative, less productive, often wrong & Bad MT is worse than no MT for specialized domains. \\
        \texttt{eng-nld-oracle-slow} & Freelance (FT) & C2 & >10   & 2-5 & 60\% & Often & Good: Productivity, repetitive work. Bad: cultural references. & More productivity at the cost of idioms and cultural factors. \\
        \texttt{eng-nld-unsup-fast}  & Freelance (FT) & C2 & 5-10 & 2-5 & 40\% & Often & Good: all. Bad: often wrong, worse quality. & PE makes you less in touch with the texts and often poorly paid. \\
        \texttt{eng-nld-unsup-avg}   & Freelance (FT) & C2 & 5-10 & 2-5 & 60\% & Sometimes & Good: Productivity, quality, repetitive work, terminology. Bad: wrong. & Practical but less effective for longer passages. \\
        \texttt{eng-nld-unsup-slow}  & Freelance (FT) & C2 & >10 & 2-5 & 40\% & Always & Good: repetitive work, productivity, terminology & Improves consistency and productivity if applied well. \\
        \texttt{eng-nld-sup-fast}    & Freelance (FT) & C2 & >10 & 5-10 & 60\% & Often & Good: repetitive work, creativity, terminology & Useful, but worries about job loss \\
        \texttt{eng-nld-sup-avg}     & Freelance (FT) & C2 & >10 & 10 & 60\% & Sometimes & Good: terminology, creativity & Useful for inspiration on better translations \\
        \texttt{eng-nld-sup-slow}    & Freelance (FT) & C1 & 5-10 & 5-10 & 80\% & Always & Good: repetitive work, productivity & Better productivity at the cost of creativity. \\
    \bottomrule[1.5pt]
    \end{tabularx}
    \begin{tabularx}{\linewidth}{llllllllllllllll}
    \toprule[1.5pt]
        \textbf{Identifier} & \textbf{Freq. Issues} & \textbf{MT quality} & \textbf{MT fluency} & \textbf{MT accuracy} & \textbf{High. accurate} & \textbf{High. useful} & \textbf{Interface clear} & \textbf{Task difficult} & \multicolumn{6}{c}{\textbf{Highlights statements}} \\
        \cmidrule(lr){10-15}
        &&&&&&&&&$\uparrow$ Speed? & $\uparrow$ Quality? & $\uparrow$ Effort? & Influence & Spot errors & $\uparrow$ Enjoy? \\
        \midrule 
        \texttt{eng-ita-nohigh-fast} & inflection,additions,omissions& 4 & 0.8 & 0.8 & - & - & 5 & 1 & - & - & - & - & - & - \\
        \texttt{eng-ita-nohigh-avg}  & multiple& 3 & 0.6 & 0.4 & - & - & 2 & 4 & - & - & - & - & - & - \\
        \texttt{eng-ita-nohigh-slow} & terminology,omissions& 3 & 0.8 & 0.8 & - & - & 1 & 5 & - & - & - & - & - & - \\
        \texttt{eng-ita-oracle-fast} & inflection,terminology& 5 & 0.4 & 0.8 & 4 & 4 & 4 & 5 & 5 & 2 & 1 & 1 & 1 & 4 \\
        \texttt{eng-ita-oracle-avg}  & syntax,terminology,omissions,no context& 3 & 0.4 & 0.6 & 2 & 1 & 2 & 3 & 1 & 1 & 4 & 1 & 1 & 1 \\
        \texttt{eng-ita-oracle-slow} & syntax,no context& 3 & 0.6 & 0.6 & 2 & 2 & 2 & 5 & 1 & 1 & 1 & 1 & 4 & 1 \\
        \texttt{eng-ita-unsup-fast}    & omissions& 3 & 0.8 & 0.6 & 3 & 2 & 4 & 5 & 3 & 3 & 3 & 2 & 2 & 2 \\
        \texttt{eng-ita-unsup-avg}     & syntax,terminology,no context& 3 & 0.6 & 0.6 & 3 & 3 & 3 & 5 & 2 & 3 & 2 & 1 & 1 & 3 \\
        \texttt{eng-ita-unsup-slow}    & syntax,inflection,terminology,omissions& 3 & 0.4 & 0.6 & 2 & 2 & 3 & 4 & 2 & 2 & 3 & 3 & 4 & 4 \\
        \texttt{eng-ita-sup-fast}  & syntax,terminology,no context& 3 & 0.4 & 0.4 & 2 & 1 & 2 & 2 & 1 & 1 & 3 & 1 & 2 & 2 \\
        \texttt{eng-ita-sup-avg}   & syntax,terminology,no context& 3 & 0.4 & 0.4 & 2 & 2 & 3 & 5 & 3 & 2 & 4 & 3 & 3 & 4 \\
        \texttt{eng-ita-sup-slow}  & syntax,terminology,omissions,no context& 3 & 0.6 & 0.6 & 2 & 2 & 1 & 2 & 2 & 1 & 1 & 4 & 4 & 1 \\
        \texttt{eng-nld-nohigh-fast} & syntax,terminology,omissions,no context& 3 & 0.2 & 0.4 & - & - & 4 & 4 & - & - & - & - & - & - \\
        \texttt{eng-nld-nohigh-avg}  & syntax,terminology,omissions,no context& 2 & 0.4 & 0.6 & - & - & 4 & 5 & - & - & - & - & - & - \\
        \texttt{eng-nld-nohigh-slow} & terminology,omissions,no context& 2 & 0.2 & 0.4 & - & - & 3 & 5 & - & - & - & - & - & - \\
        \texttt{eng-nld-oracle-fast} & syntax,inflection,terminology& 3 & 0.6 & 0.6 & 2 & 1 & 3 & 2 & 2 & 2 & 2 & 1 & 1 & 1 \\
        \texttt{eng-nld-oracle-avg}  & syntax& 3 & 0.8 & 0.6 & 4 & 3 & 3 & 4 & 3 & 3 & 3 & 3 & 2 & 3 \\
        \texttt{eng-nld-oracle-slow} & syntax,terminology& 3 & 0.6 & 0.4 & 3 & 1 & 3 & 4 & 1 & 1 & 1 & 1 & 1 & 3 \\
        \texttt{eng-nld-unsup-fast}    & terminology,additions,omissions& 3 & 0.6 & 0.8 & 3 & 2 & 4 & 4 & 1 & 3 & 1 & 1 & 2 & 1 \\
        \texttt{eng-nld-unsup-avg}     & multiple & 3 & 0.6 & 0.6 & 4 & 3 & 2 & 4 & 3 & 3 & 4 & 3 & 2 & 3 \\
        \texttt{eng-nld-unsup-slow}    & syntax,terminology,omissions & 1 & 0.4 & 0.4 & 2 & 4 & 1 & 4 & 4 & 4 & 3 & 2 & 2 & 3 \\
        \texttt{eng-nld-sup-fast}  & terminology,omissions,no context& 3 & 0.6 & 0.4 & 2 & 2 & 3 & 5 & 1 & 1 & 5 & 3 & 1 & 1 \\
        \texttt{eng-nld-sup-avg}   & syntax,additions,no context& 3 & 0.4 & 0.6 & 2 & 2 & 2 & 4 & 1 & 1 & 1 & 1 & 2 & 3 \\
        \texttt{eng-nld-sup-slow}  & multiple & 5 & 0.8 & 1   & 4 & 3 & 2 & 5 & 3 & 3 & 2 & 2 & 2 & 4 \\
    \bottomrule[1.5pt]
    \end{tabularx}
    \caption{\textbf{Top:} Sample of pre-task questionnaire results. \textbf{Bottom:} Sample of post-task questionnaire results. YoE = years of experience. Post-task statements use a 1--Strongly disagree to 5--Strongly agree scale.}
    \label{tab:questionnaire-all}
\end{sidewaystable*}

\begin{table*}[ht]
\centering
\small
\begin{tabular}{r@{\hspace{1mm}}lccccccccc}
\toprule[1.5pt]
\multicolumn{2}{c}{\multirow{2}{*}{\textbf{Modalities}}} & \multicolumn{3}{c}{\textbf{\EnIt}} & \multicolumn{3}{c}{\textbf{\EnNl}} & \multicolumn{3}{c}{\textbf{Both}} \\
\cmidrule(lr){3-5}
\cmidrule(lr){6-8}
\cmidrule(lr){9-11}
& & \textbf{Bio} & \textbf{Social} & \textbf{Both} & \textbf{Bio} & \textbf{Social} & \textbf{Both} & \textbf{Bio} & \textbf{Social} & \textbf{Both} \\
\midrule
\multirow{2}{*}{\Oracle and} & \SupervisedShort & 0.17 & 0.32 & 0.25 & \textbf{0.38} & 0.29 & 0.34 & 0.26 & 0.29 & 0.29 \\
& \UnsupervisedShort & 0.14 & 0.30 & 0.20 & \textbf{0.31} & 0.27 & 0.28 & 0.22 & 0.29 & 0.24 \\
\midrule
\multirow{2}{*}{\Supervised and} & \OracleShort & 0.19 & \textbf{0.31} & 0.26 & 0.30 & 0.26 & 0.29 & 0.24 & 0.29 & 0.28 \\
& \UnsupervisedShort & 0.19 & \textbf{0.33} & 0.25 & 0.28 & 0.24 & 0.25 & 0.24 & 0.29 & 0.25 \\
\midrule
\multirow{2}{*}{\Unsupervised and} & \OracleShort & 0.22 & 0.32 & 0.27 & \textbf{0.35} & 0.30 & 0.33 & 0.28 & 0.31 & 0.30 \\
& \SupervisedShort & 0.22 & 0.37 & 0.30 & \textbf{0.39} & 0.27 & 0.33 & 0.30 & 0.31 & 0.32 \\
\bottomrule[1.5pt]
\end{tabular}
\caption{Average highlight agreement proportion between different modalities across language pairs and domains (\Cref{sec:highlights-edits}). Scores are normalized to account for the relative frequency of highlight modalities compared to the mean highlight frequency for the current language and domain combination.}
\label{tab:highlight_agreement}
\end{table*}

\begin{figure}[htbp]
\includegraphics[width=\linewidth]{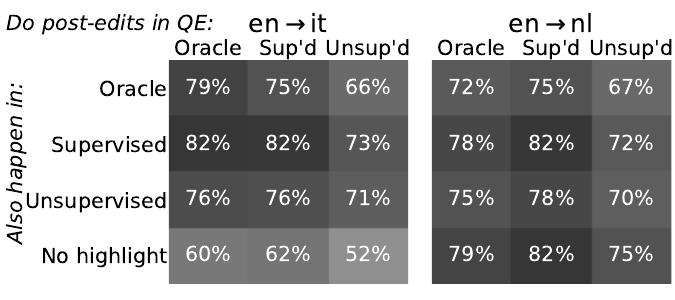}
\caption{Post-editing agreement across various modalities (\Cref{sec:highlights-edits}). Results are averaged across all translator pairs for the two modalities ($n = 3$ intra-modality, $n=9$ inter-modality for every language) and all segments.}
\label{fig:edit_agreement}
\end{figure}

\begin{figure}[htbp]
\center
\includegraphics[width=0.8\linewidth]{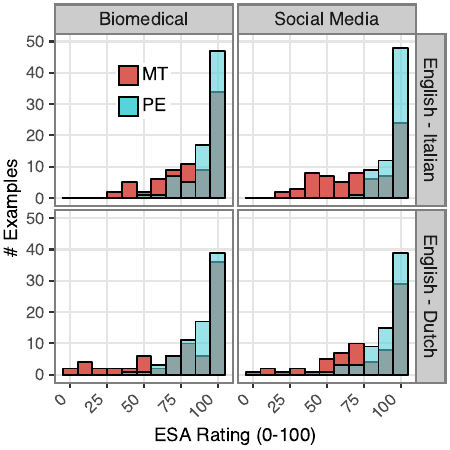}
\caption{ESA ratings for MT outputs and post-edits across domains and translation directions.}
\label{fig:esa_domains_langs}
\end{figure}

\begin{table*}[t]
\setlength{\tabcolsep}{4pt}
\small
\centering
\begin{tabular}{cl|ccc|cccc}
    \toprule[1.5pt]
    \multirow{2}{*}{\textbf{\# Doc.-Seg.}} & \multirow{2}{*}{\textbf{Error Type}} & \multicolumn{3}{c}{\textbf{Has Highlight}} & \multicolumn{4}{c}{\textbf{\% Post-edited}} \\
    \cmidrule(lr){3-5}
    \cmidrule(lr){6-9}
    & & \OracleShort & \UnsupervisedShort & \SupervisedShort & \NoHighlightShort & \OracleShort & \UnsupervisedShort & \SupervisedShort \\
    \midrule
    1-8  & Wrong number              & \NldOnly & \Both & \Both    &  67 &  \textbf{83} &  \textbf{83} &  \textbf{83} \\
    13-6 & Remove negation           & \ItaOnly & \Both & \Both    &  \textbf{50} &  33 &  33 &  \textbf{50} \\
    16-3 & Title literal translation & \Both    & \Both & \Both    &  83 &  \textbf{100} &  \textbf{100} & \textbf{100} \\
    20-1 & Wrong acronym             & \NldOnly & \Both & \ItaOnly &   0 &   \textbf{33} &   \textbf{33} &  \textbf{33} \\
    20-7 & Wrong acronym (1)         & \Neither & \Both & \Neither &   0 &   \textbf{58} &   50 &  25 \\
    20-7 & Wrong acronym (2)         & \NldOnly & \Both & \ItaOnly &   0 &   \textbf{58} &   50 &  25 \\
    22-1 & Name literal translation  & \Both    & \Both & \Both    &  50 &   50 &   \textbf{83} &  67 \\
    23-4 & Addition                  & \NldOnly & \Both & \Neither & \textbf{100} & \textbf{100}  &   83 &  50 \\
    31-2 & Wrong acronym             & \NldOnly & \Both & \Neither &  17 &  \textbf{33}  &  17  &  \textbf{33} \\
    34-7 & Numbers swapped           & \NldOnly & \Both & \NldOnly &  17 &  50  &  33  &  \textbf{67} \\
    37-4 & Verb polarity inverted    & \Both    & \Both & \Both    &  67 &  \textbf{83}  &  67  &  \textbf{83} \\
    43-5 & Wrong name                & \Both    & \Both & \Both    &  50 &  \textbf{83}  &  67  &  \textbf{83} \\
    48-5 & Wrong term                & \NldOnly & \Both & \NldOnly &  67 &  50  &  \textbf{83}  &  \textbf{83} \\
    \midrule
        & \textbf{Total}             & 65 & \textbf{100} & 62 & 44 & \textbf{63} & 60 & 60 \\
    \bottomrule[1.5pt]
\end{tabular}
\caption{Highlighting and post-editing statistics for manual critical errors (\Cref{sec:data}). Labels in \textbf{Has Highlight} columns indicate whether the error was highlighted in \Both, only one (\ItaOnly or \NldOnly) or \Neither directions. Total scores represent the percentage of detected errors (13 errors, 6 editors per highlight modality).}
\label{tab:critical-errors}
\end{table*}

\begin{table*}[t]
    \setlength{\tabcolsep}{4pt}
    \small
    \centering
    \begin{tabular}{ll|cc|ccc|ccc|c}
        \toprule[1.5pt]
\bf Domain & \bf Modality & $\mathbf{P(H)}$& $\mathbf{P(E)}$  & $\mathbf{P(E|H)}$ & $\mathbf{P(E|\neg H)}$ & $\mathbf{\Lambda_H(E)}$ & $\mathbf{P(H|E)}$ & $\mathbf{P(H|\neg E)}$ & $\mathbf{\Lambda_E(H)}$ & $\mathbf{F1_H}$\\
\midrule
\rowcolor{lightgray} \multicolumn{11}{c}{\textbf{\EnIt}}   \\
\multirow{5}{*}{Biomed.} & \textbf{Random}    &  .12 &  -   &  -  /.02 &  -  /.02 &  -  /1.0 &  -  /.11 &  -  /.13 &  -  /0.8 &  -  /.03 \\ 
                         & \NoHighlightShort  &  -   &  .02 &  -       &  -       &  -       &  -       &  -       &  -       &  -       \\ 
                         & \OracleShort       &  .08 &  .07 &  .26/.08 &  .05/.02 &  5.2/4.0 &  .30/.26 &  .06/.08 &  5.0/3.2 &  .28/.12 \\  
                         & \UnsupervisedShort &  .16 &  .10 &  .18/.06 &  .08/.02 &  2.2/3.0 &  .29/.36 &  .14/.15 &  2.0/2.4 &  .22/.10 \\  
                         & \SupervisedShort   &  .11 &  .12 &  .18/.05 &  .11/.02 &  1.6/2.5 &  .16/.23 &  .10/.10 &  1.6/2.3 &  .17/.08 \\
\midrule
\multirow{5}{*}{Social}  & \textbf{Random}    &  .20 &  -   &  -  /.09 &  -  /.09 &  -  /1.0 &  -  /.21 &  -  /.20 &  -  /1.0 &  -  /.13 \\
                         & \NoHighlightShort  &  -   &  .09 &  -       &  -       &  -       &  -       &  -       &  -       &  -       \\ 
                         & \OracleShort       &  .25 &  .20 &  .42/.23 &  .13/.04 &  3.2/5.7 &  .52/.66 &  .18/.21 &  2.8/3.1 &  .46/.34 \\ 
                         & \UnsupervisedShort &  .17 &  .18 &  .35/.19 &  .14/.07 &  2.5/2.7 &  .33/.37 &  .14/.15 &  2.3/2.4 &  .34/.25 \\ 
                         & \SupervisedShort   &  .15 &  .21 &  .38/.23 &  .18/.06 &  2.1/3.8 &  .27/.39 &  .11/.12 &  2.4/3.2 &  .32/.29 \\
\midrule
\rowcolor{lightgray} \multicolumn{11}{c}{\textbf{\EnNl}}   \\
\multirow{5}{*}{Biomed.} & \textbf{Random}    &  .17 &  -   &  -  /.12 &  -  /.10 &  -  /1.2 &  -  /.19 &  -  /.17 &  -  /1.1 &  -  /.15 \\ 
                         & \NoHighlightShort  &  -   &  .10 &  -       &  -       &  -       &  -       &  -       &  -       &  -       \\ 
                         & \OracleShort       &  .21 &  .08 &  .21/.20 &  .05/.08 &  4.2/2.5 &  .52/.41 &  .18/.18 &  2.8/2.2 &  .30/.27 \\  
                         & \UnsupervisedShort &  .23 &  .09 &  .17/.17 &  .07/.08 &  2.4/2.1 &  .43/.38 &  .21/.21 &  2.0/1.8 &  .24/.23 \\  
                         & \SupervisedShort   &  .12 &  .08 &  .20/.21 &  .06/.09 &  3.3/2.3 &  .30/.25 &  .11/.11 &  2.7/2.2 &  .24/.23 \\ 
\midrule
\multirow{5}{*}{Social}  & \textbf{Random}    &  .16 &  -   &  -  /.22 &  -  /.19 &  -  /1.1 &  -  /.19 &  -  /.16 &  -  /1.1 &  -  /.17 \\ 
                         & \NoHighlightShort  &  -   &  .19 &  -       &  -       &  -       &  -       &  -       &  -       &  -       \\ 
                         & \OracleShort       &  .19 &  .12 &  .33/.39 &  .07/.15 &  4.7/2.6 &  .54/.39 &  .15/.15 &  3.6/2.6 &  .41/.39 \\ 
                         & \UnsupervisedShort &  .15 &  .13 &  .25/.33 &  .11/.17 &  2.2/1.9 &  .30/.26 &  .13/.12 &  2.3/2.1 &  .27/.29 \\ 
                         & \SupervisedShort   &  .12 &  .10 &  .30/.36 &  .08/.17 &  3.7/2.1 &  .36/.23 &  .10/.10 &  3.6/2.3 &  .33/.28 \\
\bottomrule[1.5pt]
\end{tabular}
\caption{Highlighting ($H$) and editing ($E$) statistics for each domain, modality and translation direction combination ($n = 3$ post-editors per combination). Values after slashes are adjusted by projecting highlights of the specified modality over edits from \NoHighlight~translators to estimate highlight-induced editing biases~(\Cref{sec:proj-highlights}). A \textbf{Random} baseline is added by projecting random highlights matching the average frequency over all modalities for specific domain and translation direction settings.}
\label{tab:edit_highlights_stats_domain_modality}
\end{table*}

\begin{table*}[t]
    \setlength{\tabcolsep}{4pt}
    \small
    \centering
    \begin{tabular}{ll|cc|ccc|ccc|c}
        \toprule[1.5pt]
\bf Domain & \bf Speed & $\mathbf{P(H)}$& $\mathbf{P(E)}$  & $\mathbf{P(E|H)}$ & $\mathbf{P(E|\neg H)}$ & $\mathbf{\Lambda_H(E)}$ & $\mathbf{P(H|E)}$ & $\mathbf{P(H|\neg E)}$ & $\mathbf{\Lambda_E(H)}$ & $\mathbf{F1_H}$\\
\midrule
\rowcolor{lightgray} \multicolumn{11}{c}{\textbf{\EnIt}}   \\
\multirow{3}{*}{Biomed.} & Fast  &  \multirow{3}{*}{.09} &  .04/.01 &  .12/.02 &  .03/.01 &  4.0/2.0 &  .30/.27 &  .08/.11 &  3.7/2.4 &  .17/.04 \\
                         & Avg.  &                       &  .10/.05 &  .27/.12 &  .09/.04 &  3.0/3.0 &  .22/.30 &  .07/.11 &  3.1/2.7 &  .24/.17 \\
                         & Slow  &                       &  .09/.02 &  .21/.04 &  .08/.01 &  2.6/4.0 &  .19/.26 &  .07/.11 &  2.7/2.3 &  .20/.07 \\
\midrule
\multirow{3}{*}{Social}  & Fast  & \multirow{3}{*}{.14}  &  .11/.07 &  .30/.20 &  .07/.04 &  4.2/5.0 &  .40/.52 &  .11/.16 &  3.6/3.2 &  .34/.29 \\ 
                         & Avg.  &                       &  .23/.14 &  .48/.32 &  .18/.10 &  2.6/3.2 &  .30/.42 &  .09/.15 &  3.3/2.8 &  .37/.36 \\ 
                         & Slow  &                       &  .17/.05 &  .39/.14 &  .14/.03 &  2.7/4.6 &  .31/.54 &  .11/.17 &  2.8/3.1 &  .35/.22 \\ 
\midrule
\rowcolor{lightgray} \multicolumn{11}{c}{\textbf{\EnNl}}   \\
\multirow{3}{*}{Biomed.} & Fast  & \multirow{3}{*}{.14}  &  .03/.02 &  .11/.05 &  .02/.01 &  5.5/5.0 &  .48/.61 &  .13/.18 &  3.6/3.3 &  .18/.09 \\
                         & Avg.  &                       &  .11/.19 &  .20/.30 &  .10/.17 &  2.0/1.7 &  .25/.29 &  .13/.16 &  1.9/1.8 &  .22/.29 \\
                         & Slow  &                       &  .12/.10 &  .26/.23 &  .10/.07 &  2.6/3.2 &  .29/.42 &  .12/.16 &  2.4/2.6 &  .27/.30 \\
\midrule
\multirow{3}{*}{Social}  & Fast  & \multirow{3}{*}{.12}  &  .06/.07 &  .19/.21 &  .04/.04 &  4.7/5.2 &  .37/.47 &  .10/.13 &  3.7/3.6 &  .25/.29 \\ 
                         & Avg.  &                       &  .17/.32 &  .32/.48 &  .15/.29 &  2.1/1.6 &  .22/.23 &  .10/.12 &  2.2/1.9 &  .26/.31 \\ 
                         & Slow  &                       &  .18/.18 &  .38/.40 &  .15/.14 &  2.5/2.8 &  .25/.34 &  .09/.11 &  2.7/3.0 &  .30/.37 \\ 
\bottomrule[1.5pt]
\end{tabular}
\caption{Highlighting ($H$) and editing ($E$) statistics for each domain, and translation direction across translator speeds ($n = 4$ post-editors per combination, regardless of highlight modality). Values after slashes are adjusted by projecting highlights of the specified modality over edits from \NoHighlight~translators to estimate highlight-induced editing biases~(\Cref{sec:proj-highlights}).}
\label{tab:edit_highlights_stats_domain_speed}
\end{table*}

\begin{table*}[t]
    \setlength{\tabcolsep}{4pt}
    \footnotesize
    \centering
    \begin{tabular}{ll|cc|cc|cc|cc|cc}
    \toprule[1.5pt]
\multirow{2}{*}{\textbf{Language}} & \multirow{2}{*}{\textbf{MQM Category}} & \multicolumn{2}{c}{\textbf{MT}} & \multicolumn{2}{c}{\textbf{\NoHighlight}} & \multicolumn{2}{c}{\textbf{\Oracle}} & \multicolumn{2}{c}{\textbf{\Unsupervised}} & \multicolumn{2}{c}{\textbf{\Supervised}} \\
\cmidrule(lr){3-4}
\cmidrule(lr){5-6}
\cmidrule(lr){7-8}
\cmidrule(lr){9-10}
\cmidrule(lr){11-12}
& & \textbf{Maj.} & \textbf{Min.} & \textbf{Maj.} & \textbf{Min.} & \textbf{Maj.} & \textbf{Min.} & \textbf{Maj.} & \textbf{Min.} & \textbf{Maj.} & \textbf{Min.} \\
\midrule
\multirow{9}{*}{\textbf{Italian}} & Accuracy - Addition 
                             & 0  & 1  & 0  & 0  & 0  & 0  & 0  & 0  & 1  & 1  \\
& Accuracy - Mistranslation  & 21 & 22 & 10 & 12 & 4  & 8  & 24 & 17 & 17 & 17 \\
& Accuracy - Inconsistency   & 2  & 4  & 1  & 3  & 2  & 2  & 1  & 3  & 0  & 2  \\
& Accuracy - Omission        & 2  & 0  & 0  & 0  & 0  & 1  & 4  & 1  & 1  & 2  \\
& Accuracy - Untranslated    & 1  & 4  & 1  & 2  & 0  & 1  & 1  & 1  & 3  & 2  \\
\cmidrule(lr){2-12}
& Style - Inconsistent Style & 0  & 0  & 0  & 0  & 0  & 0  & 0  & 0  & 0  & 0  \\
& Style - Readability        & 17 & 25 & 5  & 30 & 0  & 12 & 4  & 34 & 1  & 29 \\
& Style - Wrong Register     & 0  & 8  & 0  & 3  & 0  & 3  & 1  & 1  & 3  & 2  \\
\cmidrule(lr){2-12}
& Linguistic - Grammar       & 6  & 15 & 2  & 16 & 0  & 5  & 3  & 12 & 2  & 12 \\
& Linguistic - Punctuation   & 1  & 13 & 0  & 9  & 0  & 3  & 1  & 6  & 0  & 3  \\
& Linguistic - Spelling      & 5  & 3  & 0  & 4  & 0  & 3  & 3  & 2  & 0  & 1  \\
\cmidrule(lr){2-12}
& \textbf{Total}             & 55 & 95 & 19 & 79 & 6  & 38 & 42 & 77 & 28 & 71 \\
\midrule
\midrule
\multirow{11}{*}{\textbf{Dutch}} & Accuracy - Addition        
                             & 0  & 1  & 0  & 2  & 0  & 3  & 0  & 2  & 0  & 1  \\
& Accuracy - Mistranslation  & 25 & 34 & 18 & 25 & 23 & 27 & 12 & 31 & 16 & 29 \\
& Accuracy - Inconsistency   & 0  & 0  & 0  & 2  & 0  & 2  & 0  & 2  & 0  & 5  \\
& Accuracy - Omission        & 3  & 1  & 1  & 1  & 2  & 1  & 1  & 1  & 4  & 2  \\
& Accuracy - Untranslated    & 4  & 4  & 1  & 1  & 1  & 4  & 1  & 3  & 0  & 2  \\
\cmidrule(lr){2-12}
& Style - Inconsistent Style & 2  & 0  & 0  & 5  & 1  & 7  & 0  & 2  & 0  & 9  \\
& Style - Readability        & 1  & 27 & 1  & 20 & 0  & 13 & 2  & 15 & 6  & 41 \\
& Style - Wrong Register     & 0  & 2  & 0  & 3  & 0  & 3  & 0  & 1  & 1  & 0  \\
\cmidrule(lr){2-12}
& Linguistic - Grammar       & 3  & 19 & 2  & 14 & 3  & 23 & 2  & 6  & 3  & 12 \\
& Linguistic - Punctuation   & 0  & 6  & 0  & 3  & 0  & 4  & 0  & 2  & 0  & 3  \\
& Linguistic - Spelling      & 1  & 1  & 1  & 1  & 2  & 1  & 0  & 1  & 0  & 0  \\
\cmidrule(lr){2-12}
& \textbf{Total}             & 39 & 95 & 24 & 77 & 32 & 88 & 18 & 66 & 30 & 104\\
\bottomrule[1.5pt]
\end{tabular}
\caption{MQM error counts averaged across $n = 3$ translators per highlight modality for every translation direction. A description of MQM categories is available in \Cref{fig:qa}.}
\label{tab:mqm-errors-full}
\end{table*}

\begin{figure}[t]
\centering
\includegraphics[width=0.8\linewidth]{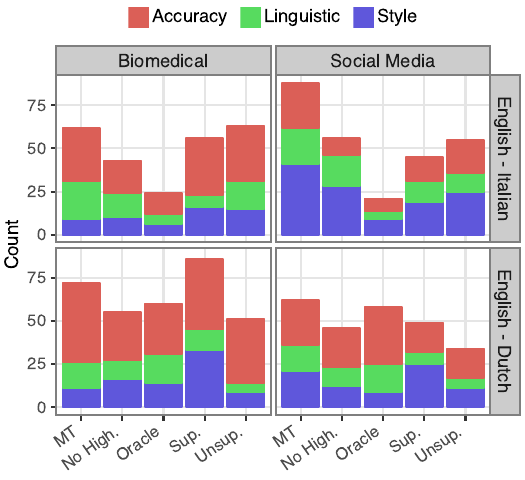}
\caption{Distribution of MQM error categories for MT and post-edits across highlight modalities for the two translation directions and domains of QE4PE.}
\vspace{-12pt}
\label{fig:quality_mqm}
\end{figure}

\begin{figure}[t]
\includegraphics[width=\linewidth]{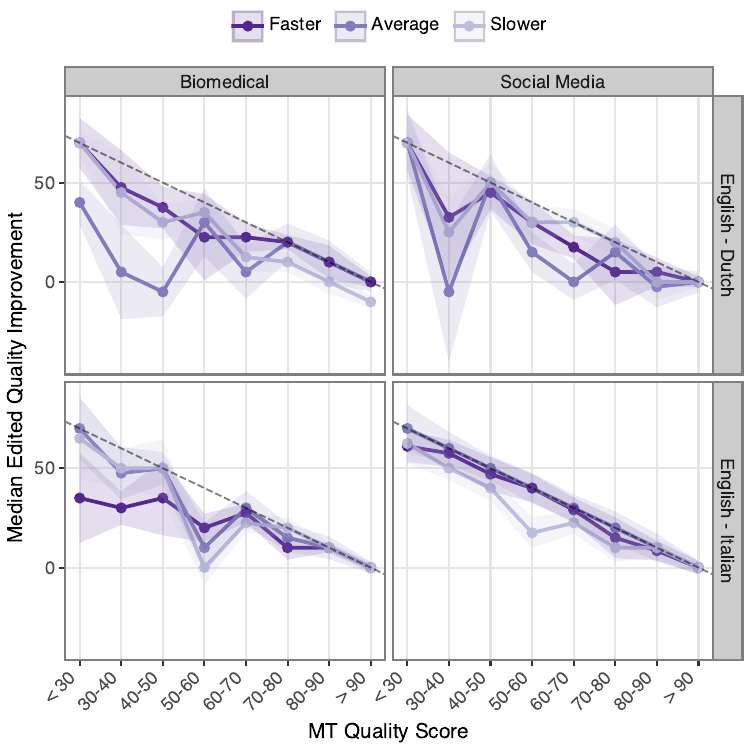}
\caption{Median ESA quality improvement following post-editing for segments at various initial MT quality levels across translators' speed groups, showing no clear quality trends across editors' productivity levels.}\vspace{-12pt}
\label{fig:quality_across_speed}
\end{figure}

\begin{table*}
\small
    \centering
    \begin{tabular}{l|cccc|cccc}
    \toprule[1.5pt]
        \multicolumn{1}{c}{\multirow{3}{*}{\textbf{Method}}} & \multicolumn{4}{c}{\textbf{DivEMT}} & \multicolumn{4}{c}{\textbf{QE4PE}} \\
    \cmidrule(lr){2-5}
    \cmidrule(lr){6-9}
        & \multicolumn{2}{c}{\textbf{En$\rightarrow$It}} & \multicolumn{2}{c}{\textbf{En$\rightarrow$Nl}} & \multicolumn{2}{c}{\textbf{En$\rightarrow$It}} & \multicolumn{2}{c}{\textbf{En$\rightarrow$Nl}} \\
    \cmidrule(lr){2-3}
    \cmidrule(lr){4-5}
    \cmidrule(lr){6-7}
    \cmidrule(lr){8-9}
        & \textbf{AP} & \textbf{AU} & \textbf{AP} & \textbf{AU} & \textbf{AP} & \textbf{AU} & \textbf{AP} & \textbf{AU} \\
    \midrule
        \textsc{Logprobs}~\citep{fomicheva-etal-2020-unsupervised}      & 0.18 & 0.18 & 0.19 & 0.19 & 0.10 & 0.09 & 0.09 & 0.09 \\
        \textsc{Logprobs}{\footnotesize~$_{\textsc{mcd var}}$}~(\citealp{fomicheva-etal-2020-unsupervised},~\UnsupervisedShort) & 0.41 & 0.41 & 0.42 & 0.42 & 0.23 & 0.23 & 0.31 & 0.31 \\
    \midrule
        XCOMET-XXL~(\citealp{guerreiro-etal-2024-xcomet},~\SupervisedShort) &  &  &  &  & 0.16 & 0.23 & 0.19 & 0.28 \\
    \midrule
        \textsc{avg.}~\Oracle~\textsc{single translator} & - & - & - & - & 0.53 & 0.73 & 0.55 & 0.75 \\
    \midrule
    
    \end{tabular}
    \caption{Average Precision (AP) and Area Under the Precision-Recall Curve (AU) between metrics and error spans derived from human post-editing. We use mBART 1-to-50~\citep{tang-etal-2021-multilingual} and NLLB 3B~\citep{nllb} respectively for DivEMT and QE4PE. For DivEMT, a single post-editor is available for computing the agreement, while for QE4PE we use consensus-based~\Oracle~highlights. For QE4PE, we report the average agreement between individual oracle post-editors and their consensus as an agreement upper bound.}
    \label{tab:unsup_selection}
\end{table*}

\begin{figure*}[htbp]
    \centering
    \hfill
    \makebox[0.32\linewidth][c]{\textbf{Pre-task}}
    \makebox[0.32\linewidth][c]{\textbf{Main task}}
    \makebox[0.32\linewidth][c]{\textbf{Post-task}}
    \includegraphics[width=1\linewidth]{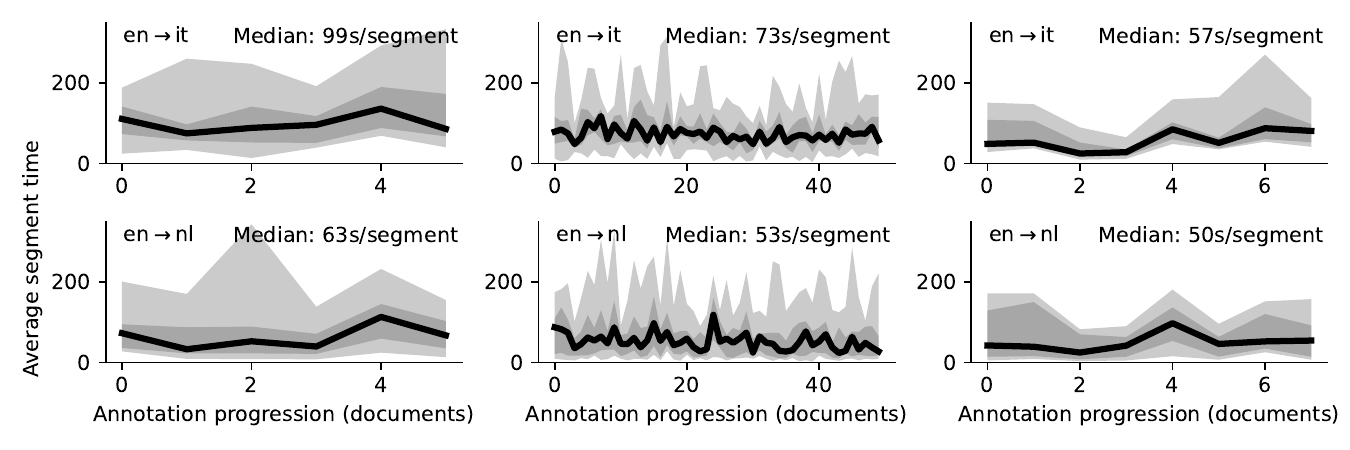}
    \caption{Segment-level post-editing time with respect to post-editor progression. Values are medians across all annotators. Light gray area is min-max values, dark gray represents 25\%-75\% quantiles. The annotators do not became considerably faster with the task progression, likely due to the simplicity of the task and the high post-editing proficiency of professional post-editors. The high variability in editing times motivates the careful group assignments performed using \textsc{Pre} task edit logs.}
    \label{fig:learning_effect_time}
\end{figure*}

\begin{figure*}[htbp]
    \centering
    \hfill
    \makebox[0.32\linewidth][c]{\textbf{Pre-task}}
    \makebox[0.32\linewidth][c]{\textbf{Main task}}
    \makebox[0.32\linewidth][c]{\textbf{Post-task}}
    \includegraphics[width=1\linewidth]{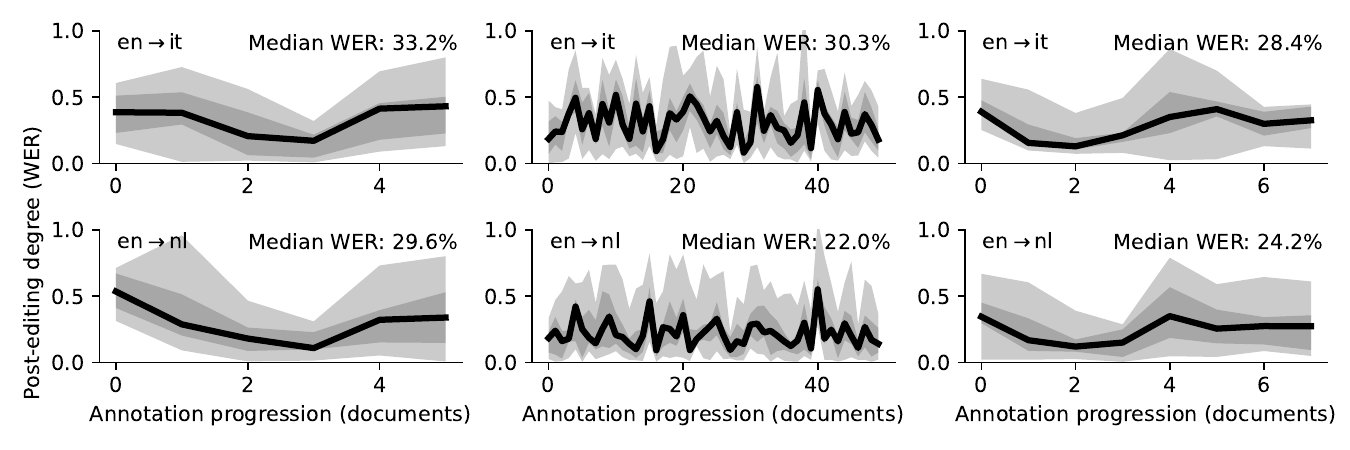}
    \caption{Editing proportion, measured by word error rate between MT and post-edited texts, with respect to post-editor progression. Values are medians across all post-editors.}
    \label{fig:learning_effect_wer}
\end{figure*}

\end{document}